\documentclass[pmlr,twocolumn,10pt]{jmlr}
\usepackage[T1]{fontenc}
\mlhtrack{proceedings}
\newif\iffinal
\finaltrue
\iffinal
  \ifmlhneedspmlr
    \jmlrvolume{346}
    \jmlryear{2026}
  \fi
  \ifmlhfindings \jmlrproceedings{}{ML4H 2026 - Findings Track}\fi
  \ifmlhdemo \jmlrproceedings{}{ML4H 2026 - Demo Track}\fi
  \jmlrworkshop{Machine Learning for Health (ML4H) 2026}
\else
  \jmlrproceedings{}{Submitted to ML4H 2026: \mlhtrackname}
  \jmlrworkshop{Machine Learning for Health (ML4H) 2026}
\fi
\usepackage{booktabs}
\usepackage{colortbl}
\usepackage{fvextra}
\usepackage[most]{tcolorbox}

\usepackage{needspace}
\definecolor{TableHeader}{HTML}{E3EEF2}
\definecolor{TableStripe}{HTML}{F3F7F8}
\definecolor{TableSummary}{HTML}{E5F1ED}
\definecolor{TableRule}{HTML}{7A9299}
\definecolor{CardInk}{HTML}{20313B}
\definecolor{SystemAccent}{HTML}{665180}
\definecolor{UserAccent}{HTML}{2A6884}
\definecolor{CodeAccent}{HTML}{566675}
\definecolor{SchemaAccent}{HTML}{956429}
\definecolor{ReplacementAccent}{HTML}{186F68}
\definecolor{RetainedAccent}{HTML}{975022}
\colorlet{PromptAccent}{SystemAccent}
\newcommand{\promptpalette}[1]{\colorlet{PromptAccent}{#1}}
\newtcblisting[auto counter]{promptbox}[1]{
  enhanced,
  unbreakable,
  listing only,
  listing engine=listings,
  colback=PromptAccent!4!white,
  colframe=PromptAccent!38!white,
  colbacktitle=PromptAccent!9!white,
  coltitle=CardInk,
  fonttitle=\sffamily\bfseries\small,
  title={\textcolor{PromptAccent}{\scriptsize PROMPT \thetcbcounter}\quad #1},
  attach boxed title to top left={xshift=3mm,yshift=-2mm},
  boxed title style={boxrule=0pt,arc=1.1mm,colback=PromptAccent!12!white,
    left=2mm,right=2mm,top=1mm,bottom=1mm},
  arc=1.2mm,
  boxrule=0.35pt,
  borderline west={1.5pt}{0pt}{PromptAccent},
  left=3mm,right=3mm,top=4mm,bottom=2mm,
  toptitle=1.5mm,bottomtitle=1.5mm,
  before skip=7pt,after skip=8pt,
  before upper={\label{promptcard:\thetcbcounter}},
  after upper={\label{promptcardend:\thetcbcounter}},
  listing options={
    basicstyle=\fontfamily{zi4}\fontsize{8}{9.3}\selectfont\color{CardInk},
    aboveskip=0pt,belowskip=0pt,
    emph={audio_id,dataset_id,language,source,target,caps,adjudications,
      cap_id,category,canonical_fact,source_quote,source_speaker,source_role,
      status,target_quote,role_status,target_role,modifier_errors,
      clinical_impact,rationale,unsupported_target_assertions,
      conversation_assessment,outcome,corruption_family,decision_status},
    emphstyle=\color{PromptAccent}\bfseries,
    breaklines=true,
    columns=fullflexible,
    keepspaces=true,
    showstringspaces=false,
    upquote=true,
    literate={“}{{``}}1 {”}{{''}}1
  }
}
\newcommand{\originalterm}[1]{\tcbox[on line,boxrule=0pt,arc=.6mm,boxsep=0pt,left=1pt,right=1pt,top=.5pt,bottom=.5pt,colback=RetainedAccent!11!white,colupper=RetainedAccent]{\textbf{#1}}}
\newcommand{\replacementterm}[1]{\tcbox[on line,boxrule=0pt,arc=.6mm,boxsep=0pt,left=1pt,right=1pt,top=.5pt,bottom=.5pt,colback=ReplacementAccent!11!white,colupper=ReplacementAccent]{\textbf{#1}}}

\title[AegisVox]{AegisVox: Factorized Privacy and Clinical Preservation for Long-Form Speech}
\author{%
\Name{Praphul Singh} \Email{praphul.singh@oracle.com}\\
\Name{Akshat Agarwal} \Email{akshat.z.agarwal@oracle.com}\\
\Name{Ratul Ghosh} \Email{ratul.ghosh@oracle.com}\\
\Name{Aashna Kanuga} \Email{aashna.kanuga@oracle.com}\\
\Name{Devashish Khatwani} \Email{devashish.khatwani@oracle.com}\\
\Name{Anurag Dwarakanath} \Email{anurag.dwarakanath@oracle.com}\\
\Name{Sri Gadde} \Email{sri.gadde@oracle.com}\\
\Name{Krishnaram Kenthapadi} \Email{krishnaram.kenthapadi@oracle.com}\\
\addr Oracle Health AI
}

\begin{document}
\maketitle

\begin{abstract}
Clinical speech reveals identity through both spoken identifiers and voice, yet its pauses, delivery, and speaker roles can also carry useful clinical evidence. Protecting that speech therefore requires more than preserving its words. We present \textbf{AegisVox}, a factorized framework that assigns consistent surrogate voices and edits identifier-bearing turns using document-consistent replacements. Across 390 public simulated or non-patient conversations totaling 63.6 hours, a matched comparison of two speech renderers separates replacement realization from retained identifiers, speaker linkage, and clinical preservation. Text-to-speech rendering realizes 535/601 planned replacements versus 453/601 for the codec-based speech editor. However, similar aggregate clinical preservation conceals more model-labeled contradictions, and conversation-wide speaker scores conceal greater edited-speech linkage under a score-direction sensitivity analysis. Fresh transcription of original audio provides a reference for interpreting the clinical measurement chain; an independent judge tests sensitivity to model-based assessment. Better replacement realization can therefore coexist with worse clinical or speaker-linkage measurements. AegisVox supports use-specific evaluation of those tradeoffs, not a guarantee of identifier removal or clinical safety.
\end{abstract}


\ifmlhneedsstatements
\paragraph*{Data and Code Availability}
We use public PriMock57, Fareez, and SIMSAMU
\citep{papadopoulos-korfiatis-etal-2022-primock57,Fareez_2022,Nun_2025},
OpenSwara target voices \citep{openswara2026}, and synthetic ASQ-PHI \citep{Weatherhead_2026}.
Appendix~\ref{app:data-access} gives revisions and redistribution restrictions.
Code is available at \url{https://github.com/spraphul/aegisvox}.
The accompanying evidence artifact documents the evaluation; source audio
remains subject to the listed access and redistribution terms.

\paragraph*{Institutional Review Board (IRB)}
Audio experiments used public simulated or non-patient recordings. The limited
five-reviewer listener study was not submitted for formal IRB review.
\fi

\section{Introduction}

\begin{figure*}[t]
\centering
\includegraphics[width=\textwidth]{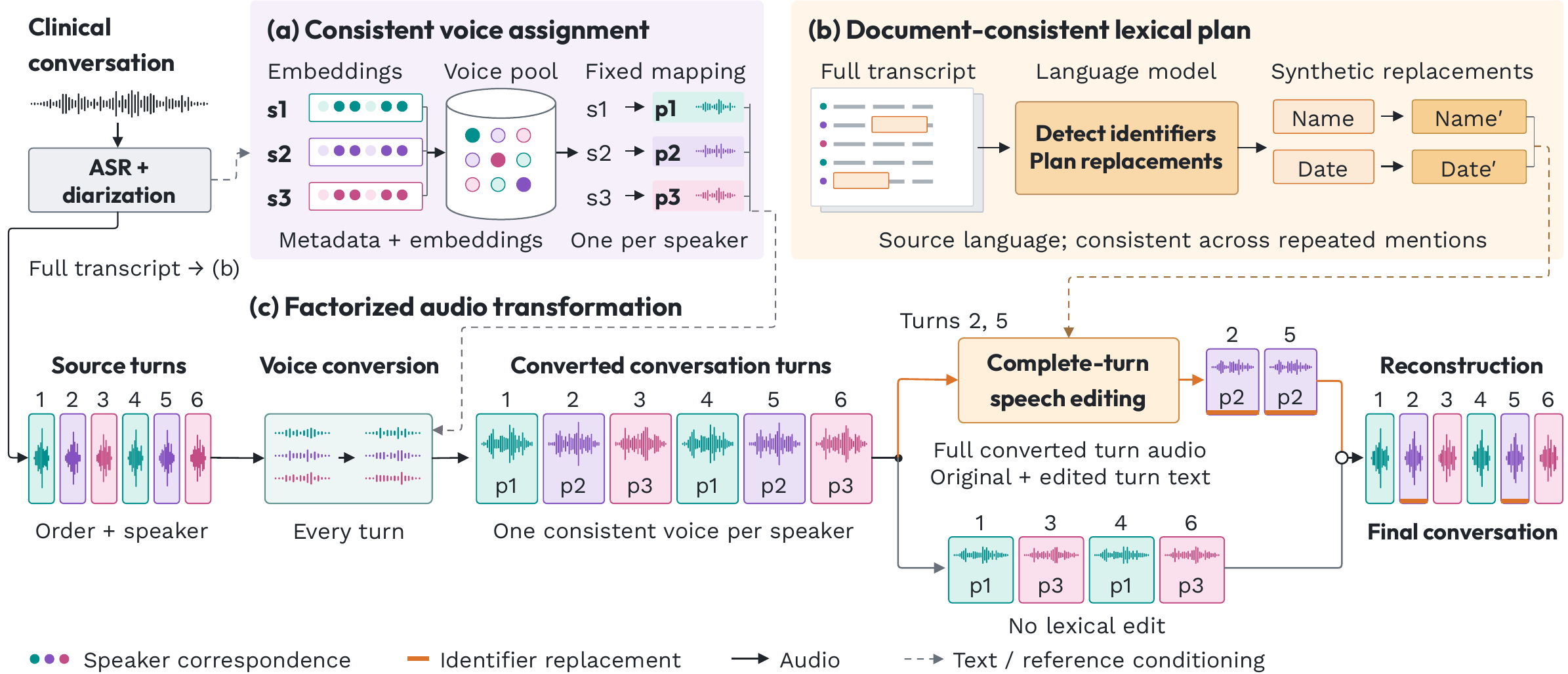}
\caption{AegisVox separates speaker identity from identifying words.
(a) Each source speaker $s_i$ receives one consistent surrogate voice $p_i$
using reference metadata and embeddings. (b) The full transcript supports
source-language replacement planning with one mapping across repeated mentions.
(c) Every turn undergoes voice conversion. Only turns with planned replacements
are rerendered as complete turns (2 and 5 in this illustration); all other
converted turns pass directly to reconstruction. Speaker colors show
correspondence, while amber marks lexical edits.}
\label{fig:method}
\end{figure*}

A clinical conversation contains more than its transcript. The words state
symptoms and decisions, while the recording preserves who spoke, when a
correction occurred, and how an answer was delivered. Pauses, prosody, accent,
and non-speech events may matter when researchers study communication or build
speech-based clinical tools. Yet that same recording exposes identity through
both spoken identifiers and vocal characteristics. Removing a name from text
leaves both its spoken form and the original voice available in the audio.

Releasing a de-identified transcript can avoid the additional risks of sharing
sound. When research requires audio, however, transformation must preserve
useful evidence without simply recreating the encounter. Muting can interrupt an utterance. Full resynthesis
can reproduce its words but cannot preserve the observed delivery by
construction. The practical question is therefore not just whether transformed
speech remains intelligible, but which identifying and clinically important
properties change together.

Prior work provides strong components for this task. Clinical
de-identification detects sensitive text or edits its spoken realization
\citep{cohn-etal-2019-audio,singh-etal-2025-redactor,kim-etal-2024-generalizing}.
Speaker anonymization transforms voice identity and measures source linkage
with automatic speaker verification (ASV)
\citep{tomashenko2020introducingvoiceprivacyinitiative,panariello2024voiceprivacy2022challengeprogress,tomashenko2024voiceprivacy2024challengeevaluation}.
Recent systems also combine speaker and lexical transformations
\citep{xiao2026anonymizationeliminationutilitypreservedspeech}.
What remains unclear is how these interventions interact in long-form clinical
conversations, where one speaker and one entity can recur across many turns.
A replacement may be spoken while an earlier mention of the original remains;
a fluent transcript may lose a negation; an overall speaker score may hide
linkability in the edited speech.

We introduce \textbf{AegisVox}, a factorized framework that changes speaker
identity and identifying words while retaining the surrounding converted
conversation. Each speaker receives a consistent surrogate voice, called a
\emph{pseudo-speaker}. A language model generates source-language synthetic
replacements for detected entities, with one mapping reused throughout the
conversation. A speech renderer then edits only the affected complete turns.
This separation supports controlled renderer comparisons and distinct tests
of detection, replacement, linkage, and clinical preservation.

Our evaluation covers 390 public simulated or non-patient clinical
conversations totaling 63.6 hours. The matched renderer comparison improves
replacement realization, but the controls reveal a more consequential finding:
similar aggregate clinical preservation does not imply similar error types,
and conversation-wide linkage does not describe every edited region.
A freshly transcribed original-audio control helps interpret the measured
clinical change without treating model judgments as clinical truth.

\noindent\textbf{Contributions.} We provide (1) a long-form speech
transformation with consistent speaker and lexical mappings; (2) a
stage-resolved evaluation that separates planned replacement, retained
identifiers, final-output linkage, and source-grounded clinical evidence; and
(3) a controlled study of two lexical renderers, with original-audio,
independent-judge, frontend, and downstream-note checks that expose where the
measures agree and where they do not.

\section{Related Work}

\paragraph{Speaker and Lexical Privacy.}
VoicePrivacy pairs attacker-oriented ASV with intelligibility and quality
measurements
\citep{tomashenko2020introducingvoiceprivacyinitiative,panariello2024voiceprivacy2022challengeprogress,tomashenko2024voiceprivacy2024challengeevaluation}.
Newer systems improve pseudo-speaker generation and utility
\citep{chen2026nouveauvoicegeneratingnovelpseudo,kuzmin2026streamvoiceanonenhancingutilityrealtime}.
Clinical text systems address a different identity surface through detection,
redaction, or consistent pseudonymization
\citep{vakili-dalianis-2022-utility,simancek-vydiswaran-2024-handling,singh-etal-2025-redactor}.
Audio de-identification must also accommodate transcription errors and realize
its text changes in speech
\citep{cohn-etal-2019-audio,abdjul-etal-2025-indonesian}.
Neither a text-detection score nor a speaker-verification score establishes
both forms of protection.

\paragraph{Speech Editing and Joint Transformation.}
Text-guided editors replace selected speech while conditioning on observed
context
\citep{tan2021editspeechtextbasedspeech,wang2022campnetcontextawaremaskprediction,borsos2022speechpaintertextconditionedspeechinpainting,peng-etal-2024-voicecraft,chen2025f5ttsfairytalerfakesfluent}.
General speech generation offers another route, but requires rechecking the
interaction it reconstructs
\citep{borsos2023audiolmlanguagemodelingapproach,shen2023naturalspeech2latentdiffusion}.
\citet{aggazzotti2026contentanonymizationprivacylongform} transform long-form
transcripts to reduce linkage through linguistic style.
\citet{xiao2026anonymizationeliminationutilitypreservedspeech} combine generated
speaker embeddings with localized identifier editing and evaluate the result
as training data. AegisVox builds on joint transformation through long-form clinical
factorization and controlled assessment of its distinct privacy and
preservation outcomes.

\paragraph{Clinical Preservation.}
Clinical note generation depends on facts, roles, temporal relations, and
supported decisions, not lexical similarity alone
\citep{nair-etal-2023-generating,ben-abacha-etal-2023-overview,yim2023acibenchnovelambientclinical}.
Word error rate (WER) does not distinguish a benign transcription variant from
a changed dose or negation. Model-based evaluation can test these distinctions
at scale, but introduces dependence on the judge
\citep{liu-etal-2023-g,zheng2023judgingllmasajudgemtbenchchatbot}.
We therefore retain separate clinical labels, test an independent judge, and
report the limits of transcript-based evidence. Our renderer comparison uses
published components within one framework; it is not a matched benchmark
against another complete privacy pipeline.

\section{AegisVox}

\subsection{Task and Consistent Mappings}
The input is a conversation waveform. Automatic speech recognition (ASR)
transcribes speech, and diarization assigns it to speakers. Together they produce
$X=\{(a_i,t_i,s_i,\tau_i)\}_{i=1}^{N}$, where $a_i$ is the waveform of turn
$i$, $t_i$ its transcript, $s_i$ its speaker, and $\tau_i$ its time interval.
Detection identifies sensitive mentions $E$ with literal transcript evidence
and turn ownership. AegisVox uses two mappings: $\pi:s\mapsto v$, from each
source speaker to a public surrogate reference voice, and $\rho:e\mapsto e'$,
from each detected entity to a synthetic replacement. Both mappings remain
fixed within a conversation.

The output is a reconstructed waveform with transformed voices and planned
lexical substitutions. A turn is \emph{affected} if it contains at least one
planned substitution. All turns undergo voice conversion; only affected turns
undergo additional lexical rendering (Figure~\ref{fig:method}). These rules
limit resynthesis, but do not by themselves guarantee correct detection,
identifier removal, or preserved delivery.

\subsection{Speaker Transformation}
The ECAPA-TDNN speaker-embedding model represents each source speaker $s$ and public
reference voice $v$ as normalized vectors $z_s$ and $z_v$ \citep{desplanques20_interspeech}.
The voice pool contains 4,062 quality-controlled OpenSwara units \citep{openswara2026} with available
language, accent, gender-presentation, and vocal-trait metadata. The FAISS similarity-search library retrieves
75 candidates per speaker. Eligible references meet duration, audio-quality,
language, and source-distance constraints. Available metadata helps avoid an
acoustically incompatible reference, but is not a privacy guarantee.

Candidates are ranked by
\begin{equation}
\begin{aligned}
D_{\cos}(s,v)&=1-\cos(z_s,z_v),\\
q(s,v)&=D_{\cos}(s,v)+2T(s,v),
\end{aligned}
\label{eq:assignment}
\end{equation}
where $T(s,v)\in[0,1]$ is a weighted match over available voice traits.
In deterministic speaker order, assignment selects the highest-ranked unused
reference satisfying source--target distance $\geq0.70$ and pairwise
target--target distance $\geq0.65$, with bounded backtracking for dead ends.
References cannot be reused within a conversation. The thresholds are untuned
heuristics: all 16 tested settings assigned all 809 conversation-speaker
profiles, and three rendered operating points changed only 14--23 assignments
without a detectable downstream contrast
(Appendix~\ref{app:retrieval-thresholds}). This is constrained ranking, not
learned global assignment or an established privacy--utility frontier.

SeedVC, a zero-shot diffusion-transformer voice converter
\citep{liu2024zeroshotvoiceconversiondiffusion}, converts every turn toward its
assigned reference. Reusing the reference maintains a consistent intended
pseudo-identity across a speaker's turns. Distinct references maintain the
intended separation between speakers, while ASV tests whether source linkage
persists after transformation.

\subsection{Lexical Planning and Complete-Turn Rendering}
GPT-5.4 receives the complete speaker-attributed transcript and generates a
structured identifier and replacement plan. The detector follows a frozen
contract based on Health Insurance Portability and Accountability Act (HIPAA)
Safe Harbor categories. Each entity receives a speakable synthetic substitute
in the source language, reused across coreferent mentions. The instructions
exclude changes to clinical facts, medication, negation, chronology, and roles.
The instruction does not verify complete detection or regulatory compliance.

Applying $\rho$ yields the complete replacement-bearing transcript $t_i'$.
For an affected turn, the renderer receives the complete SeedVC-converted
audio, its transcript $t_i$, and $t_i'$. We compare two implementations:
\emph{AegisVox (VoiceCraft)}, using a pretrained neural-codec speech editor
\citep{peng-etal-2024-voicecraft}, and \emph{AegisVox (F5-TTS)}, using a
flow-matching text-to-speech model conditioned on the reference turn
\citep{chen2025f5ttsfairytalerfakesfluent}. Each produces an edited complete
turn, not an isolated replacement word. The framework is renderer-agnostic;
these are two implementations of the same factorization.

\begin{table*}[t]
\arrayrulecolor{TableRule}
\centering\small\setlength{\tabcolsep}{3pt}
\caption{Public evaluation population and comparison roles. The original-audio
control changes no speech; all transformation conditions share the source
recordings. EN and FR denote English and French.}
\label{tab:evaluation-design}
\begin{tabular}{llrrrl}
\toprule
\rowcolor{TableHeader}
\textbf{Dataset} & \textbf{Language} & \textbf{Conversations} & \textbf{Hours} & \textbf{Median [IQR] (s)} & \textbf{Setting}\\
\midrule
PriMock57-mixed & EN & 57 & 8.6 & 556 [456--617] & Doctor--patient consultations\\
\rowcolor{TableStripe}
Fareez & EN & 272 & 51.8 & 672 [572--778] & Respiratory interviews\\
SIMSAMU & FR & 61 & 3.2 & 187 [152--234] & Simulated emergency calls\\
\rowcolor{TableSummary}
Total & EN/FR & 390 & 63.6 & 609 [471--736] & Public clinical audio\\
\bottomrule
\end{tabular}\par\medskip
\begin{tabular}{>{\raggedright\arraybackslash}p{.23\textwidth}>{\raggedright\arraybackslash}p{.31\textwidth}>{\raggedright\arraybackslash}p{.37\textwidth}}
\toprule
\rowcolor{TableHeader}
\textbf{Condition} & \textbf{Audio treatment} & \textbf{Comparison role}\\
\midrule
Original audio & None; fresh ASR and adjudication & Reference for the clinical measurement chain\\
SeedVC only & Consistent speaker conversion & Shared speaker-path control\\
AegisVox (VoiceCraft) & SeedVC + complete-turn lexical editing & Matched lexical-renderer comparison\\
AegisVox (F5-TTS) & SeedVC + complete-turn lexical rendering & Matched lexical-renderer comparison\\
Localized redaction & SeedVC + local deletion & Deletion boundary, not replacement synthesis\\
\bottomrule
\end{tabular}
\end{table*}

\subsection{Reconstruction and Measurement Boundaries}
Rendered turns replace the corresponding converted turns. The remaining
SeedVC-converted turns, gaps, and speaker ordering supply the reconstruction
context. Inserted durations may shift later timestamps; the validated lexical
replay concatenates these segments rather than enforcing original turn length. Appendix~\ref{app:implementation}
gives model and interface details, and Appendix~\ref{app:operations} describes
execution and failure accounting.

We distinguish the generated-turn benchmark from evaluation of reconstructed
conversations. The former scores every matched edit intention, retaining
failed or uncertain outcomes without selecting a successful fallback. The
latter measures saved conversation reconstructions, including 16 regions
that retain converted speech after VoiceCraft generation failure. The focused
benchmark tests one substitution at a time; its intentions are not the composed
regions counted in final reconstruction (Appendix~\ref{app:retained-input}).
A score on a generated turn is not proof that every identifier was removed
from the complete conversation.

\begin{figure}[tb]
\begin{tcolorbox}[
  enhanced,colback=white,colframe=CardInk!20!white,
  boxrule=.35pt,arc=1.2mm,
  borderline west={1.5pt}{0pt}{ReplacementAccent},
  left=2.5mm,right=2.5mm,top=1.5mm,bottom=1.5mm,
  colbacktitle=ReplacementAccent!7!white,coltitle=CardInk,
  fonttitle=\sffamily\bfseries\small,
  toptitle=1.5mm,bottomtitle=1.5mm,
  title={Replacement and Retention Can Coexist},
  attach boxed title to top left={xshift=2mm,yshift=-2mm},
  boxed title style={boxrule=0pt,arc=1.1mm,colback=ReplacementAccent!10!white,
    left=1.5mm,right=1.5mm,top=1mm,bottom=1mm},top=4mm]
\small\raggedright\color{CardInk}
{\sffamily\bfseries Source:}\par\nobreak
``your name. \originalterm{Doris}. So \originalterm{Doris}, \ldots''
\par\smallskip
{\sffamily\bfseries Intended text:}\par\nobreak
``your name. \replacementterm{Nina}. So \originalterm{Doris}, \ldots''
\par\smallskip
{\sffamily\bfseries Edited-audio ASR:}\par\nobreak
``your name. \replacementterm{Nina}. So \originalterm{Doris}, \ldots''
\par\smallskip
\textcolor{CardInk!20!white}{\rule{\linewidth}{.35pt}}
\par\smallskip
The substitute is realized, but the original remains. This observed public-data
example explains why the two outcomes need separate labels. The intended text
is reconstructed from saved planning evidence, not a captured model request
(Appendix~\ref{app:retained-input}).
\end{tcolorbox}
\end{figure}

\section{Evaluation Design}

\subsection{Population and Controlled Comparisons}
The study uses PriMock57, Fareez, and SIMSAMU
\citep{papadopoulos-korfiatis-etal-2022-primock57,Fareez_2022,Nun_2025}
(Table~\ref{tab:evaluation-design}). The recordings are public simulated or
non-patient conversations, not a sample of real patient encounters. The
primary frontend is faster-whisper large-v3 ASR
\citep{pmlr-v202-radford23a} with pyannote 3.1 diarization
\citep{Bredin_2020}. The two renderer conditions share recordings, frontend,
pseudo-speaker assignments, SeedVC outputs, planned edit intentions, and
comparison rules. Localized redaction supplies a separate deletion boundary.
Dataset access, population strata, and condition definitions appear in
Appendices~\ref{app:population}--\ref{app:conditions}.

\subsection{Separate Privacy Outcomes}
\paragraph{Detection and Realized Replacements.}
We test text detection on all 1,051 synthetic ASQ-PHI queries
\citep{Weatherhead_2026}. The primary exact typed endpoint covers 2,147 gold
elements in the intersection of the benchmark ontology and our frozen
identifier contract; the full ontology contains 2,973 elements.
This independent text test does not estimate identifier recall in the 390
source recordings.

The renderer benchmark contains 601 matched \emph{edit intentions} from 208
conversations. For each intention, GPT-5.4 examines transcripts from two ASR
models applied to the generated complete turn. It separately judges whether
the synthetic replacement is realized and whether the original identifier is
retained, allowing spelling or phonetic variants. Either judgment can be
uncertain, and both can be positive. The denominator includes failures; it is
not the number of unique turns or an exhaustive inventory of source PHI.
Exact protocols appear in Appendices~\ref{app:phi-detector},
\ref{app:clinical-results}, and~\ref{app:prompts}.

\paragraph{Linkage to the Source Speaker.}
The final-waveform test enrolls ECAPA-TDNN on original speech from the same
recording, then tests sampled speech from each saved reconstructed output.
A conversation-speaker profile is one assigned speaker within one recording,
not necessarily a unique person across the corpus. The primary trials pair
809 profiles with their own transformed speech and with other speakers in the
same conversation, yielding 809 positive and 916 negative trials. This avoids
assuming that different recordings contain different people.

Each profile pools at most ten three-second windows. Equal error rate (EER) is measured where false matches and missed matches
balance. We report both cosine score directions and the lower EER as a
score-orientation sensitivity analysis:
the direction is selected using evaluation labels, not held-out calibration.
Lower EER means easier linkage under this test. This is not an attacker trained
on converted speech. A second analysis restricts the samples to planned
inserted-turn intervals in the final output. Historical shared-stage ECAPA and
x-vector tests \citep{Snyder_2018} use different sampling and impostors; they
are reported separately in Appendix~\ref{app:asv}, not assigned to each final
renderer condition.

\subsection{Clinical Evidence and Controls}
GPT-5.4 extracts 25,778 evidence-linked \emph{clinical atomic propositions}
(CAPs) from the source ASR transcripts. A CAP is one checkable fact with its
relevant modifiers, such as ``the patient denies a penicillin allergy.'' The
inventory is frozen across conditions. The judge seeks literal evidence in
each condition's transcript and assigns preserved, partial, missing,
contradicted, or uncertain. Strict preservation counts only preserved.
Contradiction is reported separately, since two systems can have similar
preservation rates but different error profiles.

The original-audio control freshly transcribes all 390 original ingested WAVs
and adjudicates them against the same frozen CAPs. It uses the same declared
ASR model and settings as the evaluator, rather than reusing saved source
text. Historical checkpoint/runtime identity and hosted-judge temporal
stability could not be fully verified. This control measures the combined
transcription and adjudication reference, not a human transcript or facts
omitted from the initial extraction (Appendix~\ref{app:original-control}).

Both GPT-5.4 and the independent Qwen3-32B judge
\citep{qwen3technicalreport} passed a controlled qualification suite before
study adjudication. Qwen then assessed 600 shared CAPs in two presentation
orders across three conditions, and all 601 editor intentions, without seeing
GPT outcomes. Qualification tests known perturbations, not agreement with
clinicians. Appendices~\ref{app:qualification} and~\ref{app:judge-sensitivity}
report its coverage and judge dependence.

A further control generates structured clinical notes from each condition's
transcript alone. The generator receives no source CAP inventory; that
inventory enters only the subsequent evaluator. This separates note-writing
inputs from evaluation references. We also retain clinical action, role,
chronology, and final-state endpoints, plus WER, timing, events, and acoustic
similarity from CLAP and HeAR
\citep{wu2024largescalecontrastivelanguageaudiopretraining,baur2024hearhealthacoustic}.
The latter compare learned audio representations, not clinical correctness or
human naturalness (Appendices~\ref{app:clinical-results}--\ref{app:acoustic}).

\begin{figure*}[t]
\centering
\includegraphics[width=\textwidth]{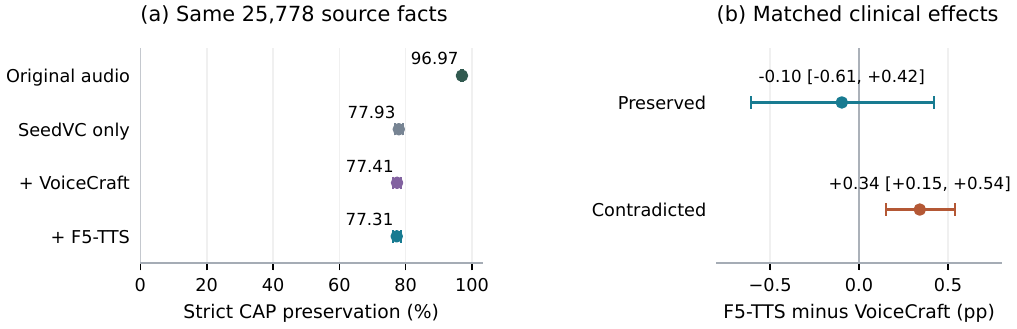}
\caption{Clinical controls and matched renderer effects on 25,778 source
CAPs from 390 conversations. (a) Fresh original-audio ASR supplies the
reference, not clinical ground truth. (b) F5-TTS minus VoiceCraft effects in
percentage points (pp), with 95\% conversation-cluster intervals. Similar
strict preservation coexists with more model-labeled contradictions.}
\label{fig:factorized-results}
\end{figure*}

\subsection{Statistical Units and Uncertainty}
Main rates are item-weighted: each edit intention or source CAP contributes
once. Paired renderer effects use 10,000 conversation-cluster bootstrap
resamples, preserving dependence among items from one recording. We also
report equal-conversation weighting. Post hoc contradiction and
retained-original tests use paired sign flips and Holm adjustment within their
two-test family. The original-control and robustness families are separate.
Final-waveform EER uses 2,000 conversation-cluster resamples; ASQ uses 2,000
query-level resamples.

Uncertain judgments do not count as verified success or verified absence.
Failed lexical cases remain in the 601-intention denominator. Invalid note
outputs remain unobserved, not zero; note contrasts use the valid matched
subset and report missingness bounds. Confidence intervals describe sampling
uncertainty, whereas bounds describe unresolved outcomes. An interval crossing
zero does not establish equivalence. Appendix~\ref{app:endpoints} defines all
units, and Appendix~\ref{app:effects} gives paired and weighting sensitivities.
Failure accounting and executable specifications appear in
Appendices~\ref{app:failures} and~\ref{app:reproducibility}.

\section{Results}

\subsection{The Original-Audio Control Explains the Reference Scale}
Fresh ASR of the original audio preserves 24,998/25,778 CAPs (96.97\%; 95\% CI
96.66--97.28), compared with 20,089/25,778 (77.93\%) after SeedVC alone
(Figure~\ref{fig:factorized-results}). The paired change is $-19.04$ percentage
points (CI $-20.29$ to $-17.83$). Both renderer rates are close to the SeedVC-only rate: 77.41\% for
VoiceCraft and 77.31\% for F5-TTS.

Most of the aggregate deficit is already present in the SeedVC-only
comparison. Similar aggregate rates do not show that the same facts failed
in each condition. It does not show
that voice conversion audibly destroys 19\% of clinical facts: transformed
speech can change ASR output, and the judge can label evidence differently.
The frozen inventory also omits facts missed by source ASR or extraction.
The control makes the absolute rates interpretable without resolving these
causes. Appendix~\ref{app:original-control} separates it from the much easier
identical-saved-text self-consistency check.

\begin{table*}[t]
\arrayrulecolor{TableRule}
\centering\small
\setlength{\tabcolsep}{4.3pt}\renewcommand{\arraystretch}{1.10}
\caption{Matched outcomes within AegisVox. Counts and item-weighted rates are
shown for each renderer. Effects are F5-TTS minus VoiceCraft, with 95\%
conversation-cluster intervals. The lexical rows cover 601 edit intentions
from 208 conversations; clinical rows cover 25,778 CAPs from 390 conversations.
Realization is conditional on a planned edit, not complete PHI removal.}
\label{tab:automatic-outcomes}
\begin{tabular}{lrrrrr}
\toprule
\rowcolor{TableHeader}
\textbf{Outcome} & \textbf{Denominator} & \textbf{VoiceCraft} & \textbf{F5-TTS} & \textbf{Effect (pp)} & \textbf{95\% CI (pp)}\\
\midrule
Replacement realized & 601 & 453 (75.37\%) & 535 (89.02\%) & $+13.64$ & $[+9.69, +17.59]$ \\
\rowcolor{TableStripe}
Original PHI retained & 601 & 21 (3.49\%) & 28 (4.66\%) & $+1.16$ & $[-0.62, +2.88]$ \\
CAP preserved & 25,778 & 19,954 (77.41\%) & 19,929 (77.31\%) & $-0.10$ & $[-0.61, +0.42]$ \\
\rowcolor{TableStripe}
CAP contradicted & 25,778 & 391 (1.52\%) & 479 (1.86\%) & $+0.34$ & $[+0.15, +0.54]$ \\
\bottomrule
\end{tabular}
\end{table*}

\subsection{More Replacements Do Not Imply Uniform Gains}
F5-TTS realizes 535/601 planned replacements (89.02\%), compared with 453/601
(75.37\%) for VoiceCraft, a gain of 13.64 points (CI 9.69--17.59;
Table~\ref{tab:automatic-outcomes}). Original PHI is nevertheless retained in
28/601 versus 21/601 edits. That difference is not statistically resolved
(Holm-adjusted $p=0.272$), but neither rate represents verified absence in all
remaining cases: original-term judgments are uncertain in 40 and 92 cases,
respectively, with one additional VoiceCraft generation failure.

Aggregate CAP preservation differs by just $-0.10$ points (CI $-0.61$ to
$0.42$). Yet F5-TTS has more model-labeled contradictions: 479/25,778 versus
391/25,778, a paired increase of 0.34 points (CI 0.15--0.54; Holm-adjusted
$p=0.0017$). This adverse secondary finding matters because strict preservation
combines several distinct failure types. Changing missing or uncertain to contradicted need not alter that rate. The labels
are not clinician-confirmed harms; they warrant inspection rather than being
hidden by the preservation average. Appendix~\ref{app:clinical-results}
reports the full outcome partition and source-assigned importance strata.

Detection remains a separate limit. On ASQ-PHI, exact typed F1 is 94.4\%
(CI 93.5--95.3) under the frozen contract intersection, but 79.9\%
(CI 78.9--80.8) under the full benchmark ontology. The difference includes a
geographic-category mismatch. Of 219 hard-negative queries, 48 elicit at least
one prediction. These are text-detection results, not recall for identifiers
in the clinical recordings; multiplying them by replacement realization would
not estimate end-to-end privacy (Appendix~\ref{app:phi-detector}).

\subsection{Final-Output Averages Can Hide Edited-Speech Linkage}
Across all-turn samples from final conversations, score-orientation EER is
0.369 for SeedVC alone, 0.365 for VoiceCraft, and 0.367 for F5-TTS. The paired
F5-TTS-minus-VoiceCraft contrast is 0.22 points (CI $-1.01$ to $0.99$), with
no detectable difference. These tests use 809 matched conversation-speaker
profiles, not 809 independently identified people.

\begin{figure}[t]
\centering
\includegraphics[width=\columnwidth]{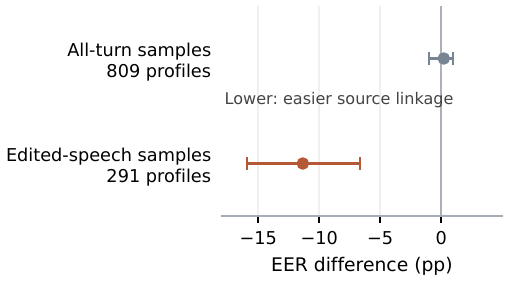}
\caption{Final-waveform linkage under evaluation-label-selected score
direction. F5-TTS minus VoiceCraft effects use all-turn samples (809 profiles,
390 conversations) or planned-edit regions (291 profiles, 204 conversations).
Bars are 95\% clustered CIs. Lower EER means easier source linkage.}
\label{fig:final-asv}
\end{figure}

Restricting samples to planned inserted-turn intervals changes the result.
On 291 matched eligible profiles from 204 conversations, F5-TTS has
11.34-point lower EER (CI $-15.91$ to $-6.69$;
Figure~\ref{fig:final-asv}). Thus, similar conversation-wide estimates do not
exclude greater source linkability in the speech being edited. Six additional
VoiceCraft profiles lack one second of eligible F5-TTS speech and are excluded
from this matched contrast, not scored as successful privacy. These findings
are sensitivity results under same-recording enrollment and selected score
direction, not a learned adaptive attack or a cross-session identification
claim (Appendix~\ref{app:asv}).

\subsection{Clinical Conclusions Depend on the Measurement Protocol}
Qwen3-32B corroborates the lexical ordering, judging 486/601 F5-TTS and 421/601
VoiceCraft replacements realized. Clinical labels nevertheless differ: Cohen's $\kappa$ is
0.253--0.340 where Qwen returns valid, agreeing labels under both presentation
orders. This unweighted comparison retains uncertainty as a separate category.
Qwen's overall weighted clinical contrast also crosses zero, but its absolute
rates and low-importance fact contrast differ from GPT-5.4. The independent
judge tests sensitivity, not clinical truth
(Appendix~\ref{app:judge-sensitivity}).

When notes are generated without the source-fact inventory, the valid matched
subset contains 314 conversations and 20,316 CAPs. F5-TTS versus VoiceCraft
strict note preservation differs by $-0.37$ points (CI $-1.30$ to $0.56$),
while model-labeled contradictions increase by 0.48 points (CI 0.24--0.72).
Only 1,440/1,560 note generations and 999/1,170 judgments are valid. These
exploratory contrasts describe observed matched cases; unresolved cases can
change the population direction (Appendix~\ref{app:notes-control}).

Five reviewers also rated 25 clinical and 25 perceptual items in a limited
blinded panel, yielding 125 ratings per panel. Its uneven coverage, including
only seven clinical and four perceptual F5-TTS items, supports descriptive
listening observations, not renderer superiority or CAP-level validation
(Appendix~\ref{app:listeners}).

\subsection{Frontend and Recording Conditions Constrain Use}
Crossing Whisper small, turbo, and large-v3 and Canary-1B-v2 with pyannote and
NeMo Sortformer yields 3,120 recording-cell evaluations
\citep{sekoyan2025canary1bv2parakeettdt06bv3efficient,pmlr-v267-park25h}.
Strict CAP preservation ranges from 0.771 to 0.807, without one backend leading
all measures. Canary has competitive matched WER, yet its interface lacks the
speaker-turn and timing links needed to localize planned edits. Observed
realization is only 2/481 intentions with NeMo and 3/596 with pyannote. These
are integration failures of the evaluated pipeline, not inherent limits of
Canary or the renderer. A transcript can be useful while its timing interface
is unusable for local speech editing (Appendix~\ref{app:frontend}).

Applying 12-dB room noise before the full pipeline reduces the
conversation-level chronology endpoint by 12.1 points (CI 6.4--17.5;
Holm-adjusted $p=0.0006$; Appendix~\ref{app:robustness}). CLAP and HeAR
similarities provide complementary acoustic evidence but do not establish
accent or naturalness invariance (Appendix~\ref{app:acoustic}). English and
French differences remain confounded with dataset, duration, and interaction
setting rather than identifying a causal language effect
(Appendix~\ref{app:language}).

\section{Discussion}

\paragraph{Different Privacy Tests Answer Different Questions.}
A spoken substitute does not prove removal of its source, and voice
conversion does not remove spoken identifiers. F5-TTS improves planned
realization, but contradictions and edited-region linkage argue against a
universal renderer choice. Neither WER nor one privacy--utility score can
express this tradeoff. The relevant balance depends on what a recipient needs
to hear and what identifying evidence an attacker can access.

\paragraph{Preserving Sound Requires More than Preserving Words.}
Complete-turn rendering supplies context without rerendering every converted
turn. It limits lexical resynthesis, not all acoustic change. The original-audio
control and frontend sensitivity show why facts, speaker attribution, timing,
and delivery require checks across the full measurement chain, not only
within the editor.

\paragraph{Review Before Release.}
Consistent pseudo-speakers can retain useful conversational roles, but matching
language and voice traits can also narrow an attacker's search. Voice-pool
knowledge, auxiliary recordings, and repeated releases could strengthen that
attacker. Restricted access, inspection of uncertain edits, and purpose-limited
release are therefore possible safeguards, not deployments validated here
(Appendix~\ref{app:governance}). No measured EER or replacement rate establishes
HIPAA compliance or a universal threshold for releasing clinical audio.

\paragraph{What Remains Unvalidated.}
The study does not provide an exhaustive independently annotated identifier
inventory for the source recordings, proposition-level clinician truth, a
learned conversion-adapted attacker, or a matched external complete pipeline.
The renderer contrast does not establish superiority over complete systems. GPT-5.4 is used for planning and primary clinical evaluation; Qwen
reduces single-family dependence but does not validate source extraction.
ASR and judging errors remain mixed with audio changes. The non-patient population, partial notes, and limited listening evidence
leave transfer to real care untested.

\section{Conclusion}
AegisVox makes long-form speech privacy a joint transformation and preservation
problem, with consistent voice assignments and document-level lexical
replacements. Its controlled evaluation shows why the outcomes must remain
separate: more planned replacements can coexist with retained identifiers,
more model-labeled contradictions, and easier edited-region linkage under the evaluated score direction.
Original-audio and final-waveform controls make these tradeoffs visible without
equating model judgments with clinical truth. Future work should pair independent clinical and identifier references with
stronger attacks on released speech.

\acks{We thank Raefer Gabriel, Kiran Rama, Vishal Vishnoi, Amitabh Saikia,
Irfan Bulu, Jangwon Kim, Neil Hauge, and Gyan Shankar for their guidance and feedback.
We thank Oracle Health AI for supporting this research and providing the
computational resources used in the experiments and analyses.}

\begingroup
\catcode95=12\relax
\catcode38=12\relax

\endgroup

\appendix
\clearpage
\section{Reporting Boundaries and Governed Use}
\label{app:governance}

The evaluation concerns public simulated or non-patient clinical conversations.
The identifier framework follows the US Department of Health and Human
Services guidance on \href{https://www.hhs.gov/hipaa/for-professionals/special-topics/de-identification/index.html}{de-identification methods}.
This framework does not replace the legal assessment required for a particular
release or establish that the transformed waveform meets that standard.
It does not establish performance on real patients, clinical safety, regulatory
compliance, guaranteed removal of protected information, or anonymity. Speaker
privacy is reported only against the evaluated automatic speaker-verification
(ASV) attackers. Lexical privacy is reported as replacement realization, retention of
the original sensitive term, and uncertainty, with each denominator stated.
Model-adjudicated clinical endpoints are not presented as human ground truth.

\begin{figure*}[t]
  \centering
  \includegraphics[width=0.94\textwidth]{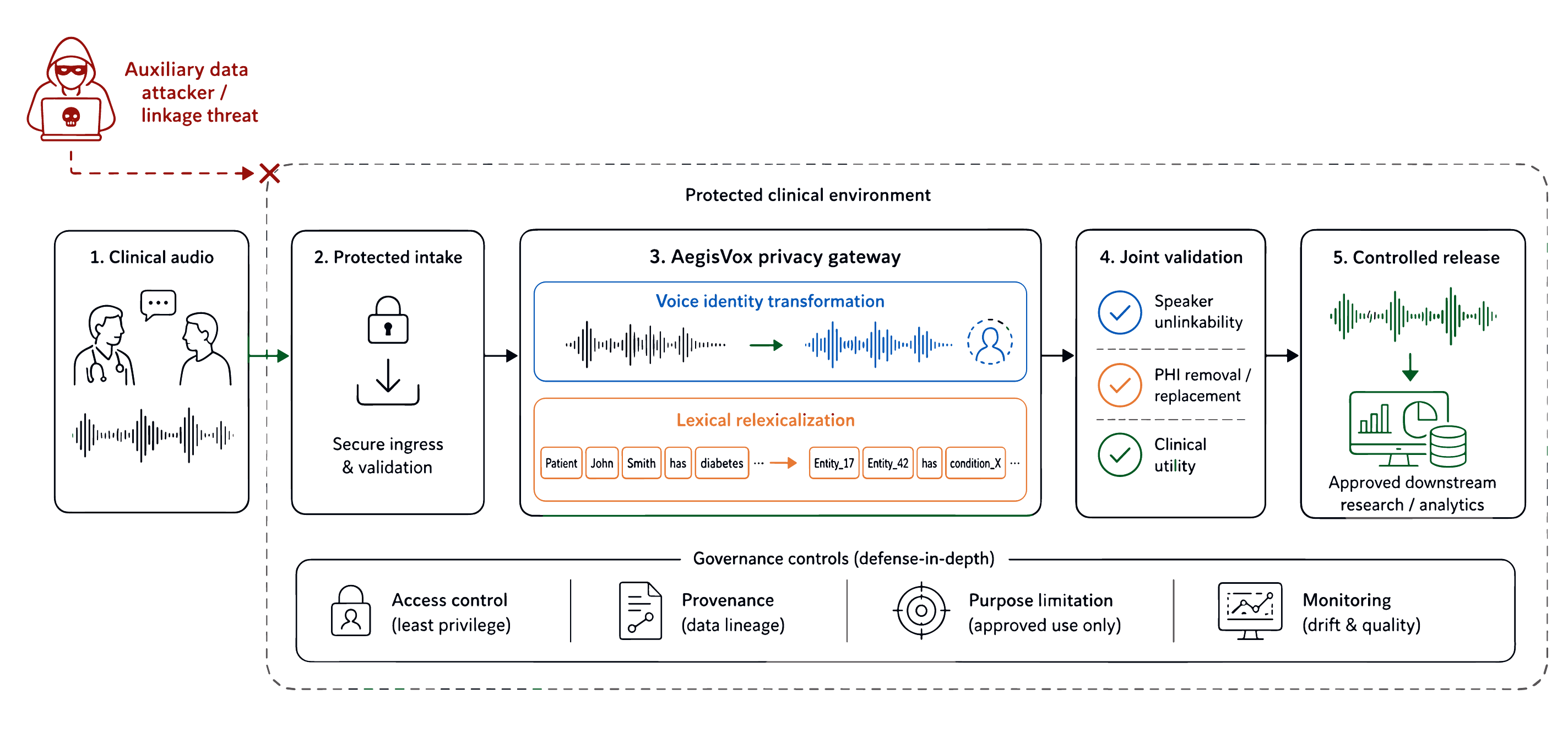}
  \caption{Governed integration and threat boundary. AegisVox transforms
  speaker identity and identifying language before controlled release, while
  access, provenance, residual-risk review, and monitoring remain outside the
  model and continue to constrain use.}
  \label{fig:governance}
\end{figure*}

Figure~\ref{fig:governance} places AegisVox inside a controlled data-release
process. The transformation is one protection layer. Source access control,
purpose limitation, provenance, residual-risk review, restricted release, and
monitoring remain necessary because an adversary may possess auxiliary voice
recordings, metadata, or repeated access to transformed outputs.

The supplement follows the same stage-resolved logic as the main paper. It
first defines the population, transformation, and controlled conditions; then
reports protected health information (PHI) detection, lexical realization,
speaker verification, and clinical
preservation under their own denominators; finally, it documents sensitivity,
failure, listener, governance, and reproducibility evidence. Results from
different statistical units are never added into a composite score.

\section{Evaluation Population and Strata}
\label{app:population}

The public evaluation contains 390 conversations and 63.6 hours of audio: 272
Fareez respiratory interviews, 57 PriMock57 consultations, and 61 SIMSAMU
emergency calls. There are 329 English and 61 French recordings. Duration
quartiles contain 98, 97, 97, and 98 recordings, with cut points near 471, 609,
and 736 seconds. Detected speaker counts are two in 302 recordings, three in 73,
and at least four in 15. These strata establish long-form and multi-speaker
coverage, but language, dataset, interaction setting, and sample size remain
confounded.

\begin{figure*}[t]
  \centering
  \includegraphics[width=0.92\textwidth]{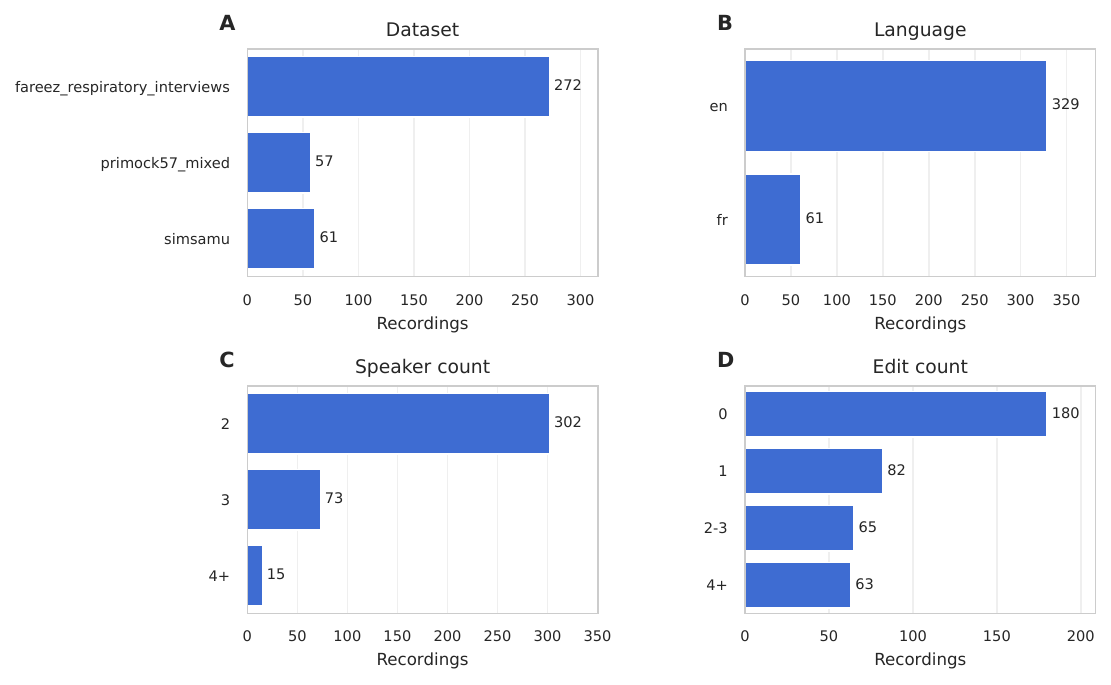}
  \caption{Public evaluation population by dataset, language, detected speaker
  count, and planned edit count. Duration strata are specified in the text.}
  \label{fig:population}
\end{figure*}

The frontend strata contain 883 timeline speaker labels, whereas the saved
ASV enrollment registry contains 809 conversation-speaker profiles: eight
recordings have one profile, 347 have two, 33 have three, and two have four.
These are different stage-specific registries, not interchangeable speaker
counts or verified unique-person identities. The timeline-label strata describe
frontend segmentation; ASV uses only its saved profile and interval registry. The edit registry
contains 611 diagnostic intentions across duration strata, while the matched
editor comparison contains 601 intentions present in both renderer manifests.
The 601-edit denominator is used for realization, original-term retention, and
uncertainty. Recording-level clinical analyses use all 390 conversations and
proposition-level analyses use the same 25,778 source clinical atomic
propositions (CAPs) per condition.
Figure~\ref{fig:population} reports the complete population composition.

The OpenSwara voice pool supplies 4,062 quality-controlled public target units.
Each stores a normalized ECAPA-TDNN embedding, waveform, duration, sample rate,
and available language, accent or geography, gender-presentation, vocal
maturity, weight, and resonance metadata. Retrieval uses these deterministic
features to seek a target that is compatible but separated from the source. A
one-to-one conversation assignment also encourages target-to-target
distinctiveness. The selected target remains fixed for that source speaker
throughout the recording. This design supports role continuity; it does not
imply protection against arbitrary auxiliary-data attacks.

\subsection{Data Access and Redistribution}
\label{app:data-access}

PriMock57 was frozen at Git revision \texttt{cd2ac707ad03} under CC BY 4.0;
the Fareez respiratory interviews use Figshare revision
\path{10.6084/m9.figshare.16550013.v1} under CC0 1.0; and SIMSAMU was
frozen at Hugging Face revision \texttt{6b639431f658} under the MIT license.
The OpenSwara pool uses revision \texttt{3fc1ab35af18}; its dataset card reports
CC BY-SA 4.0, but the evaluated snapshot does not expose uniform per-voice
upstream provenance. The accompanying artifact therefore distributes exact
manifests, identifiers, revisions, code, hashes, and aggregate evidence, but no
source, reference, transformed, or listener audio. ASQ-PHI is a fully synthetic
text benchmark released under the MIT license. Researchers obtain each public
corpus from its cited source under the corresponding access terms. The released
manifests record the complete revision identifiers.

\section{Implementation Details}
\label{app:implementation}

\paragraph{Frontend.}
The primary frontend uses faster-whisper large-v3 for matched source and target
transcription and pyannote 3.1 for speaker-attributed turns. The sensitivity
matrix additionally evaluates Whisper small, turbo, and large-v3 plus
NVIDIA Canary-1B-v2, crossed with pyannote 3.1 and NeMo Sortformer. Downstream
settings remain fixed within this matrix so that differences describe frontend
propagation rather than a reconfigured transformation.

\paragraph{Speaker Profiling and Assignment.}
Speaker turns are pooled into source profiles. FAISS inner-product search over
normalized ECAPA embeddings retrieves 75 candidates per source speaker.
Eligibility requires at least 4.5 seconds of 16-kHz audio, excludes child-like
targets, and applies language and native-English compatibility where relevant.
Reranking combines source-target distance with compatible voice traits. The
conversation-level assignment enforces one target per source speaker, no target
reuse within a conversation, and pairwise distinctiveness among assigned
targets. The assignment is made once and reused across all turns.

\paragraph{Identity Conversion.}
SeedVC converts each speaker-attributed turn toward its retrieved target voice.
The converter aims to preserve the linguistic input while changing voice
identity; its measured preservation and linkage are evaluated separately. Because all reportable
editor conditions share this conversion path, SeedVC speaker-verification
results must not be attributed to VoiceCraft or F5-TTS.

\paragraph{Replacement Planning.}
GPT-5.4 receives the complete speaker-attributed transcript, detected language,
and the conversation-level entity inventory. It emits a structured edit plan
that maps each supported sensitive entity to a synthetic replacement in the
source language. A selected replacement is reused across coreferent mentions.
The planner is restricted to identifying information and must not alter
symptoms, diagnoses, medications, doses, allergies, negation, chronology,
speaker roles, actions, or decisions.

\paragraph{Lexical Rendering and Reconstruction.}
The primary AegisVox instantiation uses the flow-matching F5-TTS model; the
matched AegisVox instantiation substitutes the pretrained, unmodified
VoiceCraft 330M neural-codec language model. Both
receive a complete converted speaker turn, its transcript, and that transcript
with the planned replacement. The resulting complete
edited turn replaces that converted turn. Unedited SeedVC turns are retained,
and reconstruction accounts for changed turn durations. The final lexical
insertion replay concatenates the inserted turns with surrounding converted
samples; it does not establish a crossfade at each insertion. No post-generation semantic rejection or favorable fallback is
inserted into the primary editor comparison.

\section{Retrieval-Threshold Sensitivity}
\label{app:retrieval-thresholds}

The source--target cutoff of 0.70 and target--target cutoff of 0.65 are
heuristic assignment constraints, not values tuned against the reported test
outcomes. We evaluated all 16 combinations of source--target minima
$\{0.60,0.65,0.70,0.75\}$ and target--target minima
$\{0.55,0.60,0.65,0.70\}$ over all 809 source profiles. Every grid cell
assigned 809/809 profiles. We then fully rendered and evaluated the
utility-oriented, balanced, and privacy-oriented settings in
Table~\ref{tab:retrieval-threshold-sensitivity} on all 390 conversations.
This is a sensitivity analysis of the declared heuristic and in-pool candidate
set. It does not test adversarial target choice or out-of-pool references.
Speaker linkage is summarized by attacker-oriented equal error rate (EER).
Utility measures include word error rate (WER) and embedding similarities from
Contrastive Language-Audio Pretraining (CLAP) and Health Acoustic
Representations (HeAR).

\begin{table*}[t]
\arrayrulecolor{TableRule}
\centering
\small
\setlength{\tabcolsep}{4.0pt}
\caption{Retrieval-threshold sensitivity over separately rendered conditions.
CAP is the recording-level mean strict proposition-preservation rate. ASV uses
the 809-positive and 653,672-impostor trial population per backend; ECAPA is
orientation-normalized so EER is at most 0.5. Absolute values belong to this
sensitivity family and are not substituted for the primary-condition results.}
\label{tab:retrieval-threshold-sensitivity}
{\small\textbf{(a) Assignment constraints and achieved distances}}\par\smallskip
\begin{tabular}{lrrrr}
\toprule
\rowcolor{TableHeader}
\textbf{Setting} & \textbf{$(D_{\mathrm{src}},D_{\mathrm{tgt}})$} & \textbf{Assigned} & \textbf{Median $D_{\mathrm{src}}$} & \textbf{Median $D_{\mathrm{tgt}}$}\\
\midrule
Utility-oriented & $(0.60,0.55)$ & 809/809 & 1.1600 & 0.8741\\
\rowcolor{TableStripe}
Balanced & $(0.70,0.65)$ & 809/809 & 1.1590 & 0.8741\\
Privacy-oriented & $(0.75,0.70)$ & 809/809 & 1.1585 & 0.8773\\
\bottomrule
\end{tabular}\par\medskip
{\small\textbf{(b) Rendered outcomes}}\par\smallskip
\begin{tabular}{lrrrrr}
\toprule
\rowcolor{TableHeader}
\textbf{Setting} & \textbf{ECAPA EER} & \textbf{x-vector EER} & \textbf{Frontend WER} & \textbf{CAP} & \textbf{CLAP / HeAR}\\
\midrule
Utility-oriented & 0.3115 & 0.3426 & 0.2314 & 0.7446 & 0.7296 / 0.9448\\
\rowcolor{TableStripe}
Balanced & 0.3103 & 0.3436 & 0.2311 & 0.7453 & 0.7297 / 0.9448\\
Privacy-oriented & 0.3140 & 0.3461 & 0.2310 & 0.7453 & 0.7303 / 0.9448\\
\bottomrule
\end{tabular}
\end{table*}

The three settings used 186 pool targets and never assigned the same target to
two speakers within one conversation. Targets may be reused across independent
conversations. Utility versus balanced changed 14/809 source assignments,
utility versus privacy changed 23/809, and balanced versus privacy changed
14/809. The small achieved-distance shifts were detectable after Holm
correction, but no downstream ASV, WER, CAP, chronology, clinical-action,
replacement-realization, CLAP, or HeAR contrast was. The evaluated cutoffs are
therefore non-binding in this pool. We retain the balanced setting for
continuity, not because it statistically dominates the alternatives. The
experiment supports robustness to these reasonable threshold choices but does
not identify a privacy--utility frontier.

\section{Controlled Conditions}
\label{app:conditions}

The four reportable conditions share the same 390 source conversations and the
same frontend. The speaker-path-only control applies only the speaker-identity
path. AegisVox (VoiceCraft) and AegisVox (F5-TTS) share the frontend,
pseudo-speaker assignment, SeedVC conversion, replacement plan, and
reconstruction, differing only in the complete-turn lexical renderer.
Localized redaction applies the same
speaker path but removes the planned lexical region without synthesizing a
replacement. Consequently, editor realization is applicable only to the two
replacement-bearing conditions. The historical ASV result measures only their shared speaker path. The
final-waveform test separately evaluates each of these three conditions.
Localized redaction begins from SeedVC-converted turns, but has no separate
final-waveform ASV result; that endpoint is unmeasured rather than zero.
The freshly transcribed original-audio condition supplies an additional
clinical measurement control without transformation.

\section{Endpoint and Statistical Protocol}
\label{app:endpoints}

Each endpoint retains its own population and stage. ASQ-PHI detection uses
1,051 synthetic queries, with 2,147 contract-intersection or 2,973 full-ontology
gold elements. Generated-turn realization and retention use 601 matched edit
intentions in 208 conversations. Final-waveform linkage uses 809
conversation-speaker profiles, with 809 positive and 916 within-conversation
negative trials for all-turn samples. Clinical preservation uses 25,778 frozen
source CAPs in 390 conversations. The frontend matrix retains all
$8\times390=3{,}120$ recording-cells. A rate from one unit cannot substitute for
coverage of another.

\paragraph{Paired Clinical and Lexical Estimates.}
Main rates are item-weighted proportions. Their paired effects use 10,000
bootstrap resamples of source conversations, retaining all items in a sampled
conversation. The random seed is 20261005. Sign-flip tests use 20,000 draws at
the same cluster level. The post hoc family of original-identifier retention
and CAP contradiction uses Holm adjustment over those two endpoints only.
Strict preservation and realization intervals are not a multiplicity-adjusted
simultaneous confidence region. Equal-conversation estimates give each
recording equal weight and are reported as a separate sensitivity analysis in
Appendix~\ref{app:effects}.

\paragraph{Controls and Other Families.}
The original-audio control also uses 10,000 conversation-cluster resamples and
20,000 sign flips. Its six-test Holm family comprises preservation and
contradiction for three transformed conditions versus original audio.
Final-waveform ASV uses 2,000 conversation-cluster resamples. ASQ-PHI uses
2,000 query-level bootstrap resamples, including the hard-negative query
proportion. The prespecified robustness families retain their separate paired
Holm analyses. These families must not be interpreted as one global error-rate
control across the paper.

\paragraph{Uncertainty Is Not Absence.}
Uncertain lexical labels and generation failures remain in the 601-intention
denominator but do not establish absence of an original identifier. The strict
CAP indicator counts only preserved labels; partial, missing, contradicted,
and uncertain remain separate in outcome tables. Invalid or absent note
judgments are unobserved, not judged missing and not a zero score. The note
comparison is restricted to valid matched cases, with no missing-at-random
assumption. Identification bounds vary unresolved outcomes and are not
confidence intervals. Structurally inapplicable measures are not assigned zero.

\section{Model-Adjudication Qualification}
\label{app:qualification}

The source-anchored clinical protocol was tested on 45 controlled cases spanning
15 families, with three cases per family. Benign controls include exact copies,
paraphrases, and identifier-only replacement. Harmful cases include critical
fact deletion, negation or uncertainty flips, medication, dose, measurement,
severity, role, plan-attribution, unsupported-insertion, and chronology errors.
The qualification matrix repeats each case under target-first ordering, rubric
permutation, prompt paraphrase, evidence removal, and context truncation.

GPT-5.4 completed 360/360 qualification judgments while satisfying the
prespecified harmful-error sensitivity, benign specificity, evidence-validity,
and fail-closed requirements. The final clinical endpoints therefore use one
qualified primary model judge. This supports a controlled model-adjudication
protocol, not equivalence to clinical experts or ground truth.

\begin{figure*}[t]
  \centering
  \includegraphics[width=0.88\textwidth]{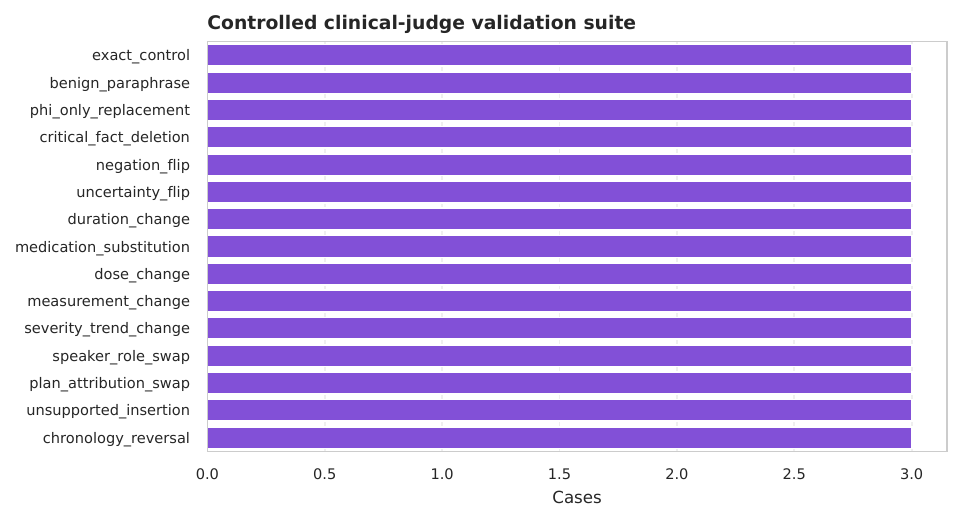}
  \caption{Controlled qualification suite for source-anchored clinical
  adjudication. Each corruption family tests whether the judge distinguishes
  preserved content from a clinically consequential change under protocol
  variants.}
  \label{fig:judge-qualification}
\end{figure*}

\subsection{Independent Open-Model Sensitivity}
\label{app:judge-sensitivity}

To test dependence on the GPT-5.4 model family used by the primary clinical
judge, Qwen3-32B
\citep{qwen3technicalreport} was frozen at Hugging Face revision
\texttt{9216db5781bf21249d13\allowbreak0ec9da846c4624c16137}, run with thinking disabled
and greedy decoding, and required to pass the unchanged qualification gates
before seeing study cases. It completed 360/360 qualification judgments. Under
the conservative source/target-order rule, consensus coverage and outcome
accuracy were 44/45, harmful sensitivity was 35/36, and benign specificity was
9/9. The sole order disagreement remained uncertain.

The qualified model then judged the same predeclared 600 source CAPs matched
across the speaker-path-only control, AegisVox (VoiceCraft), and AegisVox
(F5-TTS). Each condition case was
evaluated source-first and target-first, giving 3,600 expected judgments;
3,526 were valid and 74 remained fail-closed. GPT-5.4 outcomes were never
included in Qwen prompts. Table~\ref{tab:qwen-cap-sensitivity} separates raw counts, weighted
strict-preservation estimates, and unweighted categorical agreement. We therefore compare whether
directional conclusions persist, but do not treat the judges as interchangeable.

\begin{table*}[t]
\arrayrulecolor{TableRule}
\centering\small\setlength{\tabcolsep}{5pt}
\caption{Independent Qwen sensitivity on 600 shared CAPs per condition.
Covered cases have valid agreeing Qwen labels under both orders. Strict raw
counts retain all 600 intentions. Weighted rates use inverse stratum
probabilities. Agreement and $\kappa$ are unweighted over covered cases and
retain GPT uncertainty as a third category; they are not binary accuracy.}
\label{tab:qwen-cap-sensitivity}
\begin{tabular}{lrrrrr}
\toprule
\rowcolor{TableHeader}
\textbf{Condition} & \textbf{Covered} & \textbf{Qwen preserved / 600} & \textbf{Weighted Qwen} & \textbf{Agreement} & \textbf{$\kappa$}\\
\midrule
SeedVC only & 522 & 493 (82.2\%) & 90.9\% & 413/522 (79.1\%) & 0.253\\
\rowcolor{TableStripe}
AegisVox (VoiceCraft) & 512 & 466 (77.7\%) & 88.8\% & 405/512 (79.1\%) & 0.314\\
AegisVox (F5-TTS) & 511 & 462 (77.0\%) & 85.7\% & 405/511 (79.3\%) & 0.340\\
\bottomrule
\end{tabular}
\end{table*}

\begin{figure*}[t]
  \centering
  \includegraphics[width=0.98\textwidth]{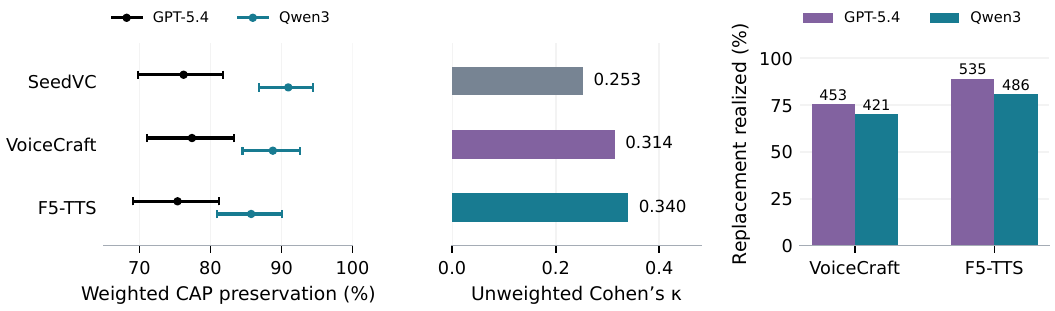}
  \caption{Independent-judge sensitivity. Left: inverse-stratum-weighted strict preservation under each judge,
  with conversation-cluster intervals. Center:
  unweighted uncertainty-retaining Cohen's $\kappa$ on covered cases. Right:
  realized replacements under each judge, with numerators over 601 intentions.
  Weighting, label agreement, and lexical ordering answer different questions.}
  \label{fig:qwen-judge-comparison}
\end{figure*}

Figure~\ref{fig:qwen-judge-comparison} separates three questions that a single
agreement number would conflate: whether population-weighted preservation estimates differ between judges,
whether they agree case by case, and
whether they recover the same lexical-renderer ordering. Weighted GPT preservation is 76.2\%, 77.4\%, and 75.4\% for SeedVC,
VoiceCraft, and F5-TTS, respectively, on the same 600-CAP sample. Qwen assigns
higher weighted preservation, while $\kappa$ shows that this is not a calibrated
substitute for GPT-5.4. Nevertheless, both judges assign higher replacement
realization to the F5-TTS instantiation than to the VoiceCraft instantiation.

Under Qwen, the paired F5-TTS-minus-VoiceCraft renderer effect within AegisVox was $-0.0305$
over all
600 CAPs (95\% audio-cluster bootstrap CI $-0.0773$ to $0.0112$). The critical
effect was $0.0193$ ($n=201$; CI $-0.0382$ to $0.0759$), the major effect was
$0.0003$ ($n=201$; CI $-0.0595$ to $0.0582$), and the minor effect was
$-0.0739$ ($n=198$; CI $-0.1483$ to $-0.0080$). The overall contrast crossed
zero under both judges, but the negative minor
stratum and systematically higher Qwen absolute rates make judge dependence a
result rather than a removable nuisance.

The same model independently judged the frozen 601-item lexical evidence.
It classified 421/601 VoiceCraft-rendered replacements as realized (70.0\%; Wilson
95\% CI 66.3--73.6) and 486/601 F5-TTS-rendered replacements as realized (80.9\%;
77.5--83.8). Original phrases were present in 29/601 and 28/601 items,
respectively. This corroborates the direction of the F5-TTS renderer's
realization gain while
showing that its absolute magnitude depends on the model judge. Both analyses
use transcript evidence, not direct listening or clinician ground truth.

\section{Independent PHI-Detector Evaluation}
\label{app:phi-detector}

The frozen text detector was evaluated on all 1,051 ASQ-PHI queries: 832
PHI-positive queries and 219 hard negatives. Gold labels were unavailable to
the model. The primary analysis uses typed exact matching over the intersection
between the frozen AegisVox contract and the benchmark ontology. It contains
2,147 gold elements and gives 2,043 true positives, 137 false positives, and
104 false negatives. Table~\ref{tab:phi-detector-overall} reports the complete
primary and sensitivity endpoints.

\begin{table*}[t]
\arrayrulecolor{TableRule}
\centering
\small
\setlength{\tabcolsep}{5pt}
\caption{Independent ASQ-PHI detector evaluation. Confidence intervals use
2,000 query-level bootstrap resamples. The full ontology includes the
prespecified geography-contract mismatch.}
\label{tab:phi-detector-overall}
\resizebox{\textwidth}{!}{%
\begin{tabular}{lrrrrrrr}
\toprule
\rowcolor{TableHeader}
\textbf{Scoring contract} & \textbf{Gold} & \textbf{TP} & \textbf{FP} & \textbf{FN} & \textbf{Precision [95\% CI]} & \textbf{Recall [95\% CI]} & \textbf{F1 [95\% CI]}\\
\midrule
Contract intersection, exact typed & 2,147 & 2,043 & 137 & 104 & 0.937 [0.926, 0.947] & 0.952 [0.942, 0.960] & 0.944 [0.935, 0.953] \\
\rowcolor{TableStripe}
Contract intersection, boundary tolerant & 2,147 & 2,124 & 56 & 23 & 0.974 [0.968, 0.981] & 0.989 [0.985, 0.993] & 0.982 [0.977, 0.986] \\
Full ASQ ontology, exact typed & 2,973 & 2,150 & 259 & 823 & 0.892 [0.880, 0.905] & 0.723 [0.714, 0.732] & 0.799 [0.789, 0.808] \\
\rowcolor{TableStripe}
Full ASQ ontology, boundary tolerant & 2,973 & 2,348 & 61 & 625 & 0.975 [0.968, 0.981] & 0.790 [0.781, 0.799] & 0.873 [0.866, 0.879] \\
\bottomrule
\end{tabular}
}
\end{table*}

The full-ontology exact F1 was 0.799 and boundary-tolerant F1 was 0.873. This
drop is not hidden as detector error under the primary contract: ASQ-PHI groups
facilities and organizations into \texttt{GEOGRAPHIC\_LOCATION}, whereas the
frozen AegisVox contract includes a location only when it is literally
geographic. On 219 hard-negative queries, 48 contained at least one predicted
entity, a query-level false-positive rate of 0.219 (95\% query-bootstrap CI
0.165--0.273). The 48 affected queries contained 49 predictions: 45 dates and
four geographic subdivisions. Most date predictions were literal years such as
2018--2022 that the frozen Safe Harbor contract treats as dates but the
benchmark labels as non-PHI in its hard-negative questions. Thus, this endpoint
captures both detector overreach and a benchmark-contract disagreement.

\begin{table*}[t]
\arrayrulecolor{TableRule}
\centering
\small
\setlength{\tabcolsep}{5pt}
\caption{Primary category results on the contract-intersection gold set.
Low-support residual categories are grouped; unique identifiers are shown
separately because their performance differs.}
\label{tab:phi-detector-category}
\begin{tabular}{lrrrrr}
\toprule
\rowcolor{TableHeader}
\textbf{Category} & \textbf{Gold} & \textbf{Exact precision} & \textbf{Exact recall} & \textbf{Exact F1} & \textbf{Boundary F1}\\
\midrule
Name & 814 & 0.937 & 0.932 & 0.935 & 0.995 \\
\rowcolor{TableStripe}
Date & 806 & 0.944 & 0.989 & 0.966 & 0.967 \\
Medical record number & 305 & 0.909 & 0.918 & 0.914 & 0.995 \\
\rowcolor{TableStripe}
Health-plan beneficiary number & 91 & 0.956 & 0.945 & 0.950 & 0.972 \\
Phone number & 45 & 1.000 & 1.000 & 1.000 & 1.000 \\
\rowcolor{TableStripe}
Social Security number & 33 & 0.943 & 1.000 & 0.971 & 0.971 \\
Email address & 31 & 1.000 & 0.968 & 0.984 & 0.984 \\
\rowcolor{TableStripe}
Unique identifier & 14 & 0.500 & 0.357 & 0.417 & 0.750 \\
Other contract categories & 8 & 1.000 & 1.000 & 1.000 & 1.000 \\
\bottomrule
\end{tabular}
\end{table*}

Unique identifiers are the weakest supported category, although only 14 gold
elements are available. Names and medical-record numbers improve sharply under
boundary-tolerant scoring, showing that many exact errors are span-boundary
disagreements rather than missed entity concepts. These results bound the
detector's supported ontology and do not establish exhaustive detection of
arbitrary identifiers in speech.

\section{Complete Clinical and Lexical Results}
\label{app:clinical-results}

The lexical chain separates an independent detector result from conditional
renderer behavior. Among the 601 edits planned for both renderers,
the AegisVox (F5-TTS) instantiation realizes 535 (89.0\%), while AegisVox
(VoiceCraft) realizes 453 (75.4\%). Original sensitive terms are retained in 28
F5-TTS-rendered edits (4.7\%) and 21 VoiceCraft-rendered edits (3.5\%); 46
F5-TTS-rendered outcomes (7.7\%) and 110 VoiceCraft-rendered outcomes (18.3\%) are
uncertain. These endpoints are adjudicated separately and may overlap, so their
counts must not be forced into a partition. The paired realization effect is
$+13.64$ percentage points (95\% conversation-cluster bootstrap CI
$9.69$ to $17.59$). Original retention differs by $+1.16$ points
(CI $-0.62$ to $2.88$; Holm-adjusted $p=0.272$). Realization uncertainty is
not retention uncertainty: the latter is 40/601 for F5-TTS and 92/601 for
VoiceCraft, with one additional VoiceCraft generation failure.

\begin{table*}[t]
\arrayrulecolor{TableRule}
\centering
\small
\setlength{\tabcolsep}{8pt}
\caption{Conditional lexical-renderer outcomes over the same 601 planned
edit intentions, assessed from generated-turn transcripts. Uncertainty refers
to realization; outcomes may overlap. Detection is reported in
Table~\ref{tab:phi-detector-overall}.}
\label{tab:lexical-renderer-chain}
\begin{tabular}{lrrrr}
\toprule
\rowcolor{TableHeader}
\textbf{Renderer} & \textbf{Planned edits} & \textbf{Replacement realized} & \textbf{Original PHI retained} & \textbf{Uncertain}\\
\midrule
VoiceCraft & 601 & 453 (75.4\%) & 21 (3.5\%) & 110 (18.3\%) \\
\rowcolor{TableStripe}
F5-TTS & 601 & 535 (89.0\%) & 28 (4.7\%) & 46 (7.7\%) \\
\bottomrule
\end{tabular}
\end{table*}

Clinical preservation is decomposed rather than summarized only by the strict
rate. Table~\ref{tab:cap-status-decomposition} shows all 25,778 proposition
judgments per condition. Strict preservation counts only \emph{preserved};
partial and uncertain cases do not pass. AegisVox (F5-TTS) preserves 19,929
propositions (77.3\%), compared with 20,089 (77.9\%) under the
speaker-path-only control. The paired F5-TTS-minus-VoiceCraft strict-CAP effect
within AegisVox is $-0.097$ percentage points
(95\% conversation-cluster CI $-0.606$ to $0.422$). This is no detectable difference under the declared test, not an
equivalence claim.

\begin{table*}[t]
\arrayrulecolor{TableRule}
\centering
\small
\setlength{\tabcolsep}{4.5pt}
\caption{Source-anchored CAP outcome decomposition. Each row contains the same
25,778 propositions; percentages are row percentages.}
\label{tab:cap-status-decomposition}
\begin{tabular}{lrrrrr}
\toprule
\rowcolor{TableHeader}
\textbf{Condition} & \textbf{Preserved} & \textbf{Partial} & \textbf{Missing} & \textbf{Contradicted} & \textbf{Uncertain}\\
\midrule
Fresh original audio & 24,998 (97.0\%) & 556 (2.2\%) & 128 (0.5\%) & 95 (0.4\%) & 1 (0.0\%) \\
\rowcolor{TableStripe}
Speaker path only & 20,089 (77.9\%) & 2,304 (8.9\%) & 2,630 (10.2\%) & 364 (1.4\%) & 391 (1.5\%) \\
AegisVox (VoiceCraft) & 19,954 (77.4\%) & 2,347 (9.1\%) & 2,663 (10.3\%) & 391 (1.5\%) & 423 (1.6\%) \\
\rowcolor{TableStripe}
AegisVox (F5-TTS) & 19,929 (77.3\%) & 2,367 (9.2\%) & 2,724 (10.6\%) & 479 (1.9\%) & 279 (1.1\%) \\
Localized redaction & 20,016 (77.6\%) & 2,225 (8.6\%) & 2,891 (11.2\%) & 387 (1.5\%) & 259 (1.0\%) \\
\bottomrule
\end{tabular}
\end{table*}

Source-extraction model labels assign 1,269 CAPs as critical, 13,486 as major,
and 11,023 as minor. These are model-generated importance strata, not clinician
annotations. Table~\ref{tab:cap-importance} shows that the aggregate rate is
not driven by one severity stratum. Most of the aggregate deficit is already present with SeedVC alone.
Similar aggregate rates do not identify which facts failed in common. They do not isolate audible
errors from transcription or judging errors, or identify the severity of a
verified clinical harm.

\begin{table*}[t]
\arrayrulecolor{TableRule}
\centering
\small
\setlength{\tabcolsep}{8pt}
\caption{Strict CAP preservation by source-assigned clinical importance.
Numerators are preserved propositions; denominators are fixed across
conditions.}
\label{tab:cap-importance}
\begin{tabular}{lrrr}
\toprule
\rowcolor{TableHeader}
\textbf{Condition} & \textbf{Critical ($n=1{,}269$)} & \textbf{Major ($n=13{,}486$)} & \textbf{Minor ($n=11{,}023$)}\\
\midrule
Speaker path only & 976 (76.9\%) & 11,044 (81.9\%) & 8,069 (73.2\%) \\
\rowcolor{TableStripe}
AegisVox (VoiceCraft) & 977 (77.0\%) & 10,951 (81.2\%) & 8,026 (72.8\%) \\
AegisVox (F5-TTS) & 959 (75.6\%) & 10,957 (81.2\%) & 8,013 (72.7\%) \\
\rowcolor{TableStripe}
Localized redaction & 989 (77.9\%) & 10,985 (81.5\%) & 8,042 (73.0\%) \\
\bottomrule
\end{tabular}
\end{table*}

\begin{figure*}[t]
  \centering
  \includegraphics[width=0.88\textwidth]{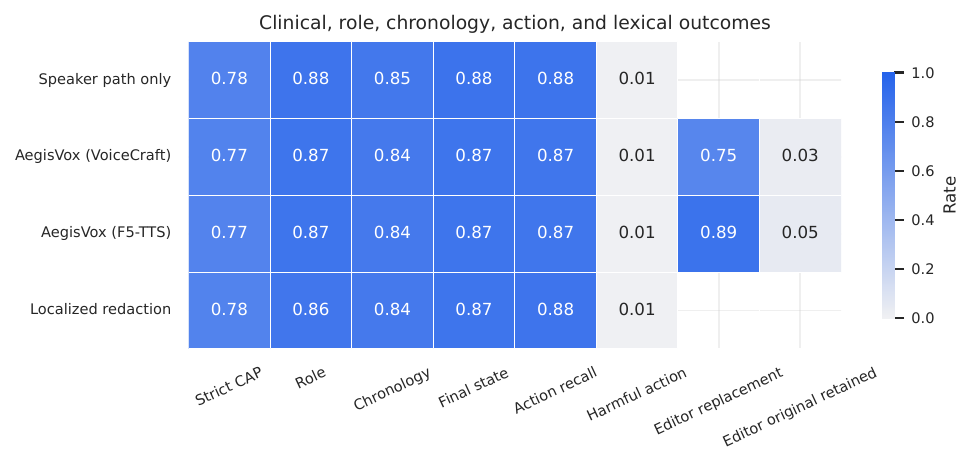}
  \caption{Complete condition-level clinical and lexical profile. Clinical
  endpoints use the 390-conversation population; editor endpoints use the
  matched 601-intention population.}
  \label{fig:clinical-lexical-full}
\end{figure*}

Figure~\ref{fig:clinical-lexical-full} aligns these clinical and lexical
outcomes without collapsing their different units. Across conditions, role
preservation ranges from 0.863 to 0.875, chronology preservation from 0.840 to
0.851, final-state preservation from 0.871 to 0.883, and action recall from
0.866 to 0.881. Harmful-action-change rates are between 0.006 and 0.007.

\subsection{Fresh Original-Audio Control}
\label{app:original-control}

All 390 normalized original ingested WAVs were transcribed afresh with
faster-whisper large-v3, CUDA float16, beam size 1, voice-activity detection,
and automatic language selection. No audio identity or lexical transformation
was applied. GPT-5.4 adjudicated these new transcripts against the existing
25,778 source CAPs; the inventory was not extracted again. All 390
adjudications were valid, with no unobserved CAPs. The original-control labels
are 24,998 preserved, 556 partial, 128 missing, 95 contradicted, and one
uncertain. Strict preservation is 96.974\% (95\% conversation-cluster CI
96.658--97.279).

The fresh original comparison gives SeedVC, VoiceCraft, and F5-TTS changes of
$-19.043$, $-19.567$, and $-19.664$ percentage points, respectively. Their
intervals are $[-20.288,-17.833]$, $[-20.830,-18.359]$, and
$[-20.923,-18.462]$. These are item-weighted paired label contrasts, not
independent estimates of audible clinical damage.

Fresh-execution receipts establish that the transcripts were newly computed;
335 differ from the saved source text and 55 are identical. Text equality
alone neither proves nor disproves fresh execution. The declared model and
settings match the evaluator, but exact historical checkpoint/runtime
identity and hosted-judge temporal stability could not be fully verified.
Source ASR and extraction omissions remain outside the frozen inventory.
A separate self-consistency check, using identical saved source text, preserves
25,775/25,778 CAPs (99.99\%); it is not the fresh original-waveform control.

\subsection{Paired Contradiction and Uncertainty}
For model-labeled contradiction, 237 CAPs are positive under both renderers,
242 under F5-TTS only, 154 under VoiceCraft only, and 25,145 under neither.
The resulting 479 versus 391 contradictions give $+0.341$ percentage points
(CI $0.150$--$0.542$), with sign-flip $p=0.000850$ and two-test
Holm-adjusted $p=0.001700$. The source importance strata describe the original
fact, not clinician-assigned severity of an output error.

If uncertain CAP labels are allowed to vary rather than assigned the strict
indicator, the preservation difference has identification bounds
$[-1.738,0.985]$ points and contradiction $[-1.300,1.424]$ points. These bounds
are not sampling intervals. They show why the reported contradiction increase
is a difference in explicit model labels, not a complete estimate of unknown
clinical outcomes.

\subsection{Retained Identifiers in Intended Inputs}
\label{app:retained-input}
Saved planning evidence allows reconstruction of intended target text for the
focused editor benchmark. An original literal is still present in 12/28
F5-TTS cases labeled retained and 12/21 VoiceCraft cases labeled retained.
This is not a corrected retention rate: every observed retained case remains
in its original denominator. Actual outgoing model requests were not saved,
so intended text cannot establish runtime compliance or the cause of each
retained identifier.

For the public PriMock example in the main text, the source begins ``your
name. Doris. So Doris, what I think we need to do is, this sounds like you
might have a chest infection.'' The reconstructed intended transcript changes
only the first occurrence to ``Nina.'' Both ASR transcripts of the F5-TTS
edited turn support ``Nina'' followed by ``Doris.'' The two lexical labels
are therefore realized replacement and retained original. The example uses
transcript evidence, not independent listening, and does not establish a
clinical error from other ASR wording differences.

The 601-intention focused benchmark and the composed final-conversation
reconstruction are separate populations. Saved final-waveform receipts cover
545 planned regions per renderer: all 545 contain generated F5-TTS turns,
whereas VoiceCraft contains 529 generated turns and 16 retained converted
regions after generation failure, across 14 conversations. The focused
VoiceCraft benchmark has one generation failure among its 601 intentions.
Neither failure count may be substituted for the other.

\section{Acoustic, Timing, and Documentation Preservation}
\label{app:acoustic}

CLAP similarity under bounded dynamic time warping (DTW) ranges from 0.725 to
0.732 and HeAR similarity from
0.945 to 0.946 across the four conditions. These representation-level measures
probe preservation of broad audio and health-acoustic structure but are not
diagnostic accuracy or human perceptual judgments. Detector-defined event
recall and precision are 0.405 and 0.240 for AegisVox (F5-TTS), 0.376 and 0.218 for
AegisVox (VoiceCraft), and 0.407 and 0.244 for the speaker-path-only control. Speech-activity changes range
from approximately $-0.047$ to $-0.059$, while speaking-rate changes range from
approximately $+5.4$ to $+8.6$ words per minute.

The earlier note comparison supplies source facts to the note-writing protocol
and compares source- and target-grounded structured notes.
Section Jaccard ranges from 0.944 to 0.960 and exact-structure agreement from
approximately 0.74 to 0.79 across generators and conditions. Comparisons are
made within note generator because generator-specific length and organization
are not valid cross-system effects.

\begin{figure*}[t]
  \centering
  \includegraphics[width=\textwidth]{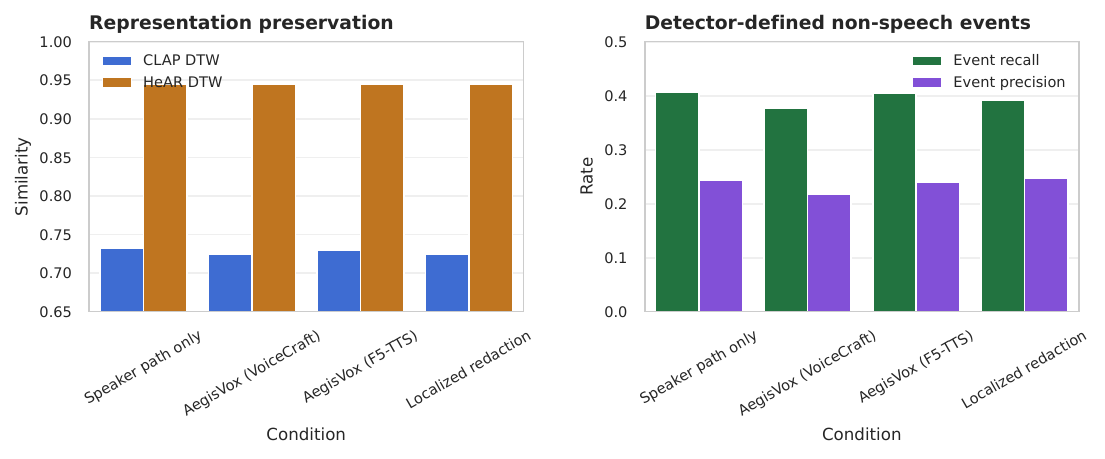}
  \caption{Acoustic representation, event, and timing preservation. CLAP and
  HeAR are embedding similarities, events are detector-defined, and timing is
  measured with the matched ASR protocol.}
  \label{fig:acoustic-preservation}
\end{figure*}

\subsection{Transcript-Only Note Generation}
\label{app:notes-control}

A separate control gives GPT-5.4 only the transcript for the current condition
and a dataset-prefixed record identifier. It receives no source CAP inventory,
replacement plan, other-condition transcript, or renderer label. Source facts
are supplied only to the later note evaluator. This tests documentation
without directly conditioning generation on the answer inventory; it does not
make GPT-generated and GPT-adjudicated notes independent clinical truth.

Of 1,560 note jobs, 1,440 yield valid notes and 120 exhaust the allowed
attempts. Of 1,170 planned target-note judgments, 999 are valid, 167 lack a
valid note pair, and four are quarantined because of duplicate unsupported-
assertion evidence. Missing jobs are not empty notes or zero preservation.
Table~\ref{tab:note-control-coverage} retains all scheduled conversations.

\begin{table*}[t]
\arrayrulecolor{TableRule}
\centering\small
\caption{Transcript-only note control. Each note condition schedules 390
conversations. Target judgments compare a valid transformed note with a valid
source note. V, Q, and B denote valid, quarantined, and blocked judgments;
invalid and missing outcomes remain unobserved.}
\label{tab:note-control-coverage}
\begin{tabular}{lrrrr}
\toprule
\rowcolor{TableHeader}
\textbf{Condition} & \textbf{Valid notes} & \textbf{Exhausted} & \textbf{Judgments (V / Q / B)} & \textbf{CAPs}\\
\midrule
Original & 362 & 28 & Not applicable & Not applicable\\
\rowcolor{TableStripe}
SeedVC only & 359 & 31 & 331 / 2 / 57 & 21,515\\
AegisVox (VoiceCraft) & 355 & 35 & 329 / 1 / 60 & 21,485\\
\rowcolor{TableStripe}
AegisVox (F5-TTS) & 364 & 26 & 339 / 1 / 50 & 22,063\\
\bottomrule
\end{tabular}
\end{table*}

The common valid F5-TTS/VoiceCraft subset has 314 conversations and 20,316
CAPs. Strict preservation is 14,597/20,316 (71.85\%) versus 14,673/20,316
(72.22\%), a difference of $-0.374$ percentage points (95\% clustered CI
$-1.302$ to $0.558$). Contradictions are 369/20,316 (1.82\%) versus
272/20,316 (1.34\%), a difference of $+0.477$ points (CI $0.239$--$0.722$).
Preservation is the primary observed-subset contrast; contradiction is
exploratory, without an adjusted significance claim. Inference uses 10,000
conversation-cluster bootstrap draws and 20,000 sign flips, seed 20261005.

Selection remains unresolved. Varying uncertain and unobserved outcomes gives
full-population difference bounds of $[-16.30,17.01]$ points for strict
preservation and $[-17.40,15.91]$ for contradiction. These are identification
bounds, not CIs, and can reverse the observed direction. Source-inclusion
judgments disagree on 307 matched CAPs. A separate conditional comparison uses
only the 19,401 CAPs judged included in both source-note assessments; its
strict-preservation contrast is $-0.505$ points (CI $-1.467$ to $0.441$).
It is not based on two unequal condition-specific denominators.

No causal numerical comparison is made with the earlier fact-conditioned
note experiment: validated coverage and protocols differ. The exact
transcript-only generation prompt and its evaluation prompt are included in
Appendix~\ref{app:prompts}.

\section{Attacker-Oriented Speaker Privacy}
\label{app:asv}

\subsection{Final-Waveform Protocol}
The final-output test first verifies the 1,170 saved reconstructed WAVs
(390 recordings in each of three conditions). Replaying the inserted generated
or retained converted regions reproduces every WAV within two PCM16
quantization steps. This verifies the file lineage, not semantic correctness.
The replay concatenates surrounding converted samples and the inserted turns;
turn duration changes shift later timestamps. It does not establish exact
preservation of original timing or crossfades at lexical insertion boundaries.

The attacker uses SpeechBrain ECAPA-TDNN at revision
\texttt{0f99f2d0ebe89ac095bc\allowbreak c5903c4dd8f72b367286}.
Enrollment comes from assigned original speech in the same recording, not
an independent session or the retrieved pseudo-speaker reference. Speech
intervals use saved pipeline assignments, not independently verified speaker
labels. Sampling takes at most ten equispaced three-second windows over the
time-ordered assigned intervals, with 50-ms edge trimming and at least one
second of usable speech. Each window embedding is unit normalized; the mean
is normalized again to form one profile.

The primary all-turn analysis has 809 conversation-speaker profiles, 809
same-profile positives, and 916 other-speaker negatives within the same
conversation. A person may recur across conversations, so cross-recording
pairs are not assumed to be independent known identities. Both score
directions are reported in Table~\ref{tab:final-asv}. The smaller EER selects
direction on evaluation labels and is a sensitivity result, not a calibrated
or trained adaptive attacker. Intervals use 2,000 conversation-cluster draws.

\begin{table*}[t]
\arrayrulecolor{TableRule}
\centering\small
\caption{Final-waveform ECAPA scores on all-turn samples: 809 positive and
916 within-conversation negative trials per condition. Lower-direction EER is
selected using evaluation labels. The interval describes resampled
conversations, not uncertainty about unseen attacker strategies.}
\label{tab:final-asv}
\begin{tabular}{lrrr}
\toprule
\rowcolor{TableHeader}
\textbf{Condition} & \textbf{High-cosine EER} & \textbf{Low-cosine EER} & \textbf{Low-cosine 95\% CI}\\
\midrule
SeedVC only & 0.6310 & 0.3690 & [0.3483, 0.3878]\\
\rowcolor{TableStripe}
AegisVox (VoiceCraft) & 0.6354 & 0.3646 & [0.3458, 0.3861]\\
AegisVox (F5-TTS) & 0.6332 & 0.3668 & [0.3469, 0.3856]\\
\bottomrule
\end{tabular}
\end{table*}

The paired all-turn effect is $+0.218$ percentage points (CI $-1.010$ to
$0.988$). The planned-edit-region analysis has 297 eligible VoiceCraft profiles
in 207 conversations and 291 F5-TTS profiles in 204 conversations. There are
512 profiles per condition without an edit interval (not applicable) and six
additional F5-TTS profiles with less than one second of eligible speech
(unknown). The matched 291-profile contrast is $-11.340$ points
(CI $-15.910$ to $-6.688$). Planned-edit regions include retained converted
audio after generation failure; they are not restricted to successful
synthesis. The unmatched group EERs, 0.4389 and 0.3230, must not be subtracted
to obtain the paired effect.

\subsection{Historical Shared-Stage Protocol}
The following separate test measures the common SeedVC stage. It uses
different sampling and cross-recording negatives, and is not a trajectory
that can be subtracted from the final-waveform estimates above. Its two
backends establish sensitivity to the verification representation, not
final-output performance of both lexical renderers.

The attacker enrolls one profile from each of the 809 original conversation-speaker profiles.
For source speaker $s$, transformed audio derived from $s$ is the positive test;
transformed audio derived from each of the other 808 source speakers forms the
impostor set. This produces 809 positive and $809\times808=653{,}672$ negative
trials per backend and condition. The attacker does not enroll on the retrieved
pseudo-speaker reference. It has access to original source-speaker audio and to
the transformed release, which is the relevant original-to-release linkage
question.

ECAPA and x-vector use the same trial identities but independent embeddings.
Because score conventions differ by backend, the informed attacker may select
the orientation that best links the source. Under this attacker-optimal
orientation, ECAPA EER is 0.304 (95\% source-bootstrap CI 0.289--0.324) and
x-vector EER is 0.345 (0.331--0.361). The ECAPA orientation is
lower-score-indicates-same-source; x-vector is
higher-score-indicates-same-source. These estimates belong only to the shared SeedVC stage. They are not the
final-waveform scores for either renderer.

\begin{table*}[t]
\arrayrulecolor{TableRule}
\centering
\small
\setlength{\tabcolsep}{6pt}
\caption{Historical shared-stage speaker-verification protocol and results. Higher
EER means harder source linkage only under the declared enrollment, trials,
and attack backend.}
\label{tab:asv-protocol}
\resizebox{\textwidth}{!}{%
\begin{tabular}{lrrrrl}
\toprule
\rowcolor{TableHeader}
\textbf{Backend} & \textbf{Source profiles} & \textbf{Positive trials} & \textbf{Impostor trials} & \textbf{EER (95\% CI)} & \textbf{Attacker orientation}\\
\midrule
SpeechBrain ECAPA & 809 & 809 & 653,672 & 0.304 (0.289--0.324) & Lower score means same source \\
\rowcolor{TableStripe}
SpeechBrain x-vector & 809 & 809 & 653,672 & 0.345 (0.331--0.361) & Higher score means same source \\
\bottomrule
\end{tabular}
}
\end{table*}

These results show residual source linkability rather than anonymous speech.
They do not model enrollment on external pseudo-speaker material, arbitrary
metadata, repeated releases, stronger future attacks, or human recognition.

\begin{figure}[t]
  \centering
  \includegraphics[width=\columnwidth]{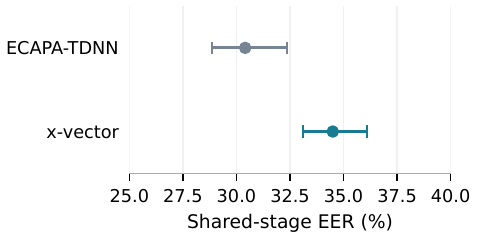}
  \caption{Historical shared-SeedVC-stage verification under two embedding
  backends, not final-renderer scores. Bars are 95\% source-bootstrap intervals.}
  \label{fig:asv-full}
\end{figure}

\section{Language and Population Strata}
\label{app:language}

The action-safe endpoint is met in 319/329 English conversations (97.0\%) and
48/61 French conversations (78.7\%). The chronology-safe endpoint is met in
186/329 English conversations (56.5\%) and 16/61 French conversations (26.2\%).
The strict conversation-level CAP no-major-error endpoint is much more severe
than proposition-level preservation because one major error causes the entire
conversation to fail. It is 5/329 for English and 2/61 for French. These
language-associated differences are not causal language effects because the
French recordings are a smaller SIMSAMU emergency-call set and differ in
setting, role structure, duration, and content.

\begin{figure*}[t]
  \centering
  \includegraphics[width=\textwidth]{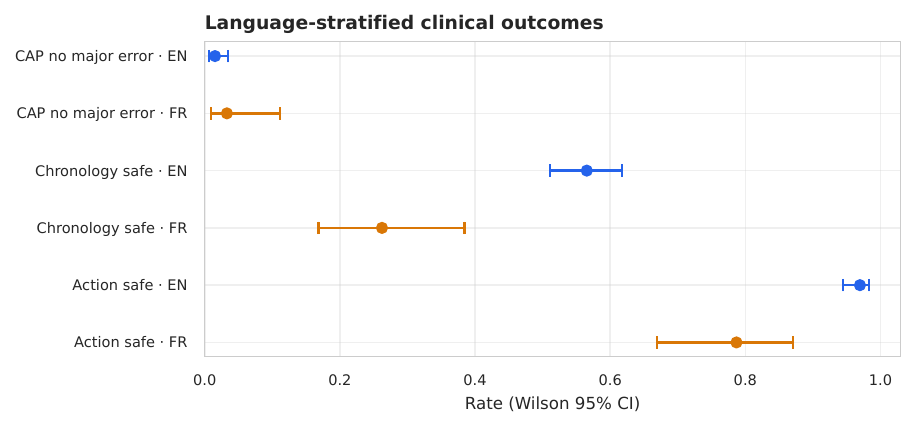}
  \caption{Language-stratified clinical outcomes with exact denominators and
  95\% intervals. Dataset and interaction setting are confounded with language.}
  \label{fig:language-strata}
\end{figure*}

\section{Frontend Propagation}
\label{app:frontend}

The full matrix contains eight ASR-by-diarization cells and 3,120
recording-cell evaluations. Every cell processes the same 390 conversations.
ASR varies among Canary-1B-v2 and Whisper small, turbo, and large-v3;
diarization varies between NeMo Sortformer and pyannote 3.1. CAP ranges from
0.771 to 0.807, action recall from 0.859 to 0.884, and cell-specific
frontend-ASR matched WER from 0.217 to 0.239. This is not the English-only MedASR
sensitivity WER; the recognizer and
population differ, so the absolute values are not comparable. The endpoints move differently across cells, demonstrating that the
frontend is part of the evaluated system rather than interchangeable
preprocessing.

The Canary lexical cells expose a different failure than transcription error.
Every planned Canary edit lacked a linked speaker turn and edit timing
(481/481 with NeMo and 596/596 with pyannote), so complete-turn lexical editing
could not be localized. The intention-to-transform analysis retained these
upstream failures and used only global final-audio ASR evidence. Consequently,
replacement realization was observed for just 2/481 items (0.42\%) with NeMo
and 3/596 (0.50\%) with pyannote, compared with 0.835--0.869 for the Whisper
cells. The similar collapse under both diarizers, despite Canary's competitive
matched WER, localizes the problem to the Canary-to-editor turn/timing contract
rather than to F5-TTS or to either diarization backend. These values therefore
measure end-to-end frontend integration failure, not the intrinsic editing
quality of Canary or F5-TTS, and they were retained rather than treated as
missing observations.

\begin{figure*}[t]
  \centering
  \includegraphics[width=0.91\textwidth]{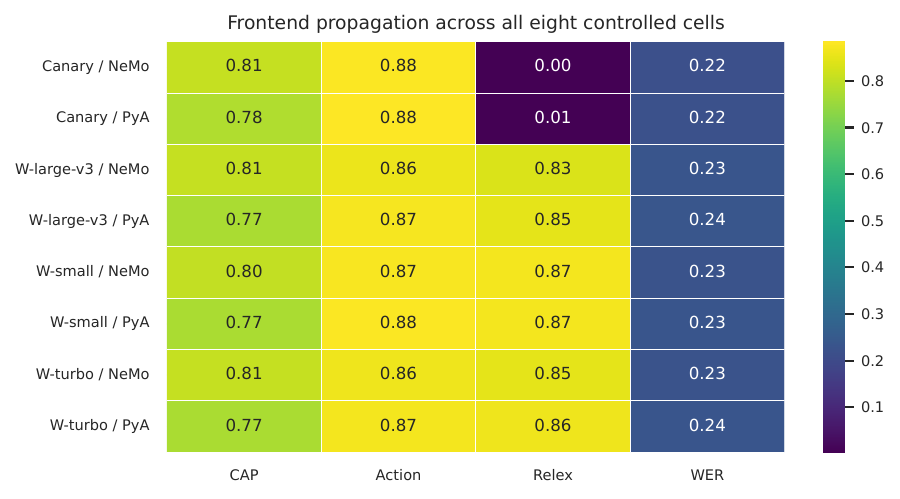}
  \caption{Propagation of ASR and diarization choices into matched content,
  clinical, action, and lexical measurements. Canary lexical cells retain
  upstream turn/timing-linkage failures in their intention-to-transform
  denominators. W denotes Whisper; PyA denotes pyannote.}
  \label{fig:frontend-full}
\end{figure*}

\section{Acoustic and Semantic Robustness}
\label{app:robustness}

Acoustic stress contributes 1,170 intention-to-transform rows, of which 1,169
are renderable and one is an explicit failure. Four semantic stress families
contribute 1,560/1,560 renderable rows. The estimand is the paired change in
conversation-level chronology safety relative to the unperturbed AegisVox
(F5-TTS) instantiation.

Room noise at a 12-dB signal-to-noise ratio (SNR) has the largest effect,
$-0.121$ (95\% CI $-0.175$ to
$-0.064$; Holm-adjusted $p=0.0006$). Telephony is $-0.031$ with an interval
crossing zero. Microphone or bandpass degradation is $-0.056$, but does not
remain significant after multiplicity adjustment. The four semantic-stress
effects range from $-0.008$ to $-0.031$ and are compatible with no change after
adjustment. The clearest measured vulnerability is therefore ordinary acoustic
degradation propagating through the frontend. For each acoustic condition, the
degradation was applied to the source waveform before rerunning the unchanged
full pipeline, including ASR and diarization; it was not applied after
transcription.

\begin{figure*}[t]
  \centering
  \includegraphics[width=\textwidth]{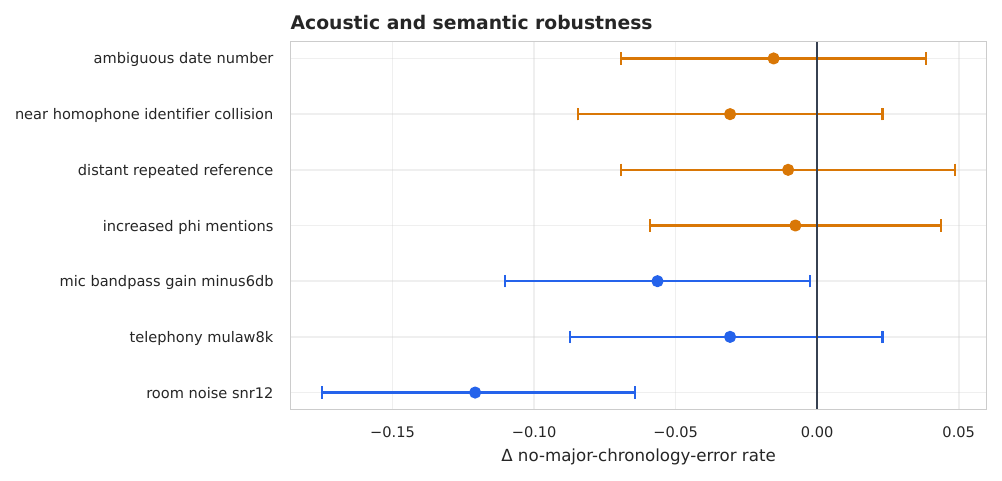}
  \caption{Paired chronology effects under acoustic and semantic stresses.
  Intervals are 95\% confidence intervals; multiplicity is controlled within
  the prespecified robustness family.}
  \label{fig:robustness-full}
\end{figure*}

\section{Limited Blinded Listener Panel}
\label{app:listeners}

Five reviewers completed a limited blinded panel containing 25 clinical and 25
perceptual items, giving 125 ratings per panel. The incomplete-block design
allocates different numbers of items to each condition, so the estimates are
descriptive and are not used for formal system superiority. Clinical ratings
cover meaning, completeness, chronology, role, and replacement quality.
Perceptual ratings cover naturalness, intelligibility, completeness, speaker
consistency, accent, and edit integration. Reviewers use a five-point scale and
may select insufficient evidence.

Mean condition ratings are high across the observed sample. For AegisVox (F5-TTS),
the item-level means are 4.43 for meaning, 4.89 for clinical completeness, 4.40
for replacement quality, 4.45 for naturalness, 4.90 for intelligibility, and
4.95 for edit integration. Because AegisVox (F5-TTS) contributes seven clinical and
four perceptual items, these measurements are a limited qualitative check rather
than a population-wide estimate.

Agreement varies by endpoint. The intraclass correlation coefficient ICC(2,1)
ranges from negative for chronology to 0.68 for intelligibility, while
ICC(2,k) is higher for most endpoints because
it describes the mean of five ratings. This variation reinforces the decision
not to promote the panel to a powered comparison or a human speaker-linkability
claim.

\section{Primary Effects and Privacy--Utility Projections}
\label{app:effects}

The item-weighted realization effect is $+13.644$ percentage points
(95\% conversation-cluster CI $9.687$--$17.590$), and the strict CAP effect is
$-0.097$ points (CI $-0.606$ to $0.422$). Under equal-conversation weighting,
the strict CAP contrast is $-0.088$ points (CI $-0.797$ to $0.593$).
The distinction matters because long conversations contribute more CAPs to the
item-weighted estimate. Neither interval establishes equivalence. Realization
intervals here use conversations as clusters, not independent edit resampling.
Speaker EER, lexical retention, and clinical outcomes are not collapsed into
one significance test.

\begin{figure*}[t]
\centering
\includegraphics[width=\textwidth]{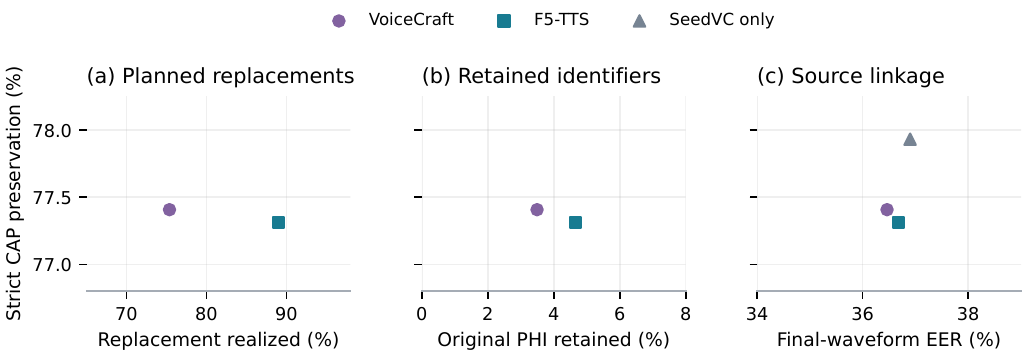}
\caption{Distinct privacy endpoints versus strict CAP preservation. Points
are descriptive condition estimates under different endpoint populations,
not a joint confidence region or a dominance test. Left and center: generated-
turn lexical judgments, with uncertainty retained in their denominators.
Right: final-waveform all-turn ECAPA EER under evaluation-label-selected score
direction. No final-waveform x-vector estimate is implied.}
\label{fig:pareto-matrix}
\end{figure*}

Figure~\ref{fig:pareto-matrix} places three distinct privacy measurements
beside item-weighted CAP preservation. Lexical realization and original
retention concern 601 intentions in 208 conversations; final-waveform linkage
concerns 809 profiles in all 390 conversations. Clinical preservation uses
25,778 CAPs in all 390 conversations. These are endpoint-level projections,
not a joint sample-level utility function or proof of Pareto dominance.

The original-retention axis counts explicit present labels, not their
complement as verified absence. Explicit absence is 487/601 (81.03\%) for
VoiceCraft and 533/601 (88.69\%) for F5-TTS; uncertain labels are 92 and 40,
with one additional VoiceCraft failure. Thus, ``one minus retained'' would
incorrectly include unknown outcomes as absence. Speaker points use the new
final-waveform ECAPA protocol; historical common-stage ECAPA or x-vector
values are not assigned to renderer outputs.

\section{Failure Taxonomy}
\label{app:failures}

Failures are assigned to the earliest measured stage that exposes them, while
their original statistical unit is retained. At detection, the contract-primary
ASQ-PHI analysis contains 104 false-negative and 137 false-positive elements;
48/219 hard-negative queries contain at least one prediction. At rendering,
replacement is not credited as realized in 148/601 VoiceCraft-rendered edits
(24.6\%) and 66/601 F5-TTS-rendered edits (11.0\%). These totals include
uncertain labels and one VoiceCraft failure, not only confirmed absence. Original sensitive terms remain in 21/601
(3.5\%) and 28/601 (4.7\%), respectively. Uncertain evidence remains visible
for 110/601 VoiceCraft-rendered and 46/601 F5-TTS-rendered edits.

At the clinical-evidence stage, AegisVox (F5-TTS) has 2,367 partial, 2,724 missing,
479 contradicted, and 279 uncertain proposition judgments among 25,778 CAPs.
Conversation- or evidence-level families additionally include harmful action
change near 0.6--0.7\%, chronology not preserved near 14.9--16.0\%, and role
not preserved near 12.5--13.7\%. Frontend integration creates a distinct
failure mode: Canary's missing turn/timing linkage prevents localization even
when its global transcript is usable. Room noise creates another by degrading
upstream evidence before editing. These heterogeneous quantities identify
different intervention targets and must not be summed into a global failure
rate.

\begin{figure*}[t]
  \centering
  \includegraphics[width=\textwidth]{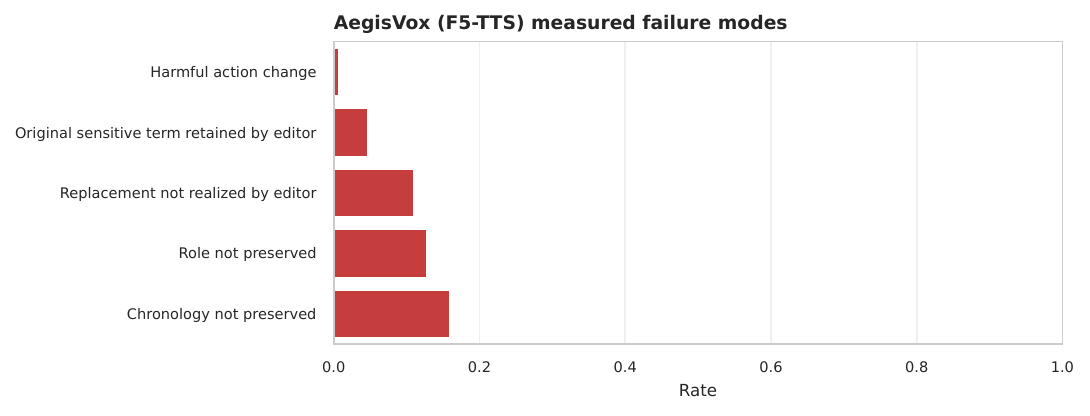}
  \caption{Outcome-linked failure taxonomy. Rates with different statistical
  units are shown separately and must not be summed into a composite failure
  score.}
  \label{fig:failure-taxonomy}
\end{figure*}

\section{Operational Lessons}
\label{app:operations}

\paragraph{Persistent and Restartable Stages.}
Long-form batches benefit from resident model processes, bounded work queues,
idempotent stage contracts, immutable recording identifiers, and durable status
artifacts. Finalization verifies the expected population, waveform duration,
hashes, and explicit failure registry. These controls prevent restart events
from silently changing the evaluated denominator.

\paragraph{Complete-Turn Editing.}
Short isolated windows often provide insufficient linguistic and speaker
context. Complete converted turns supply the editor with the source waveform,
complete source text, and complete replacement-bearing target text. This unit
also permits the generated output waveform to be bound directly to
reconstruction and evaluated independently of later policy decisions.

\paragraph{Document and Language Consistency.}
Replacement planning uses the complete transcript and source language. One
synthetic mapping is reused across repeated references. This is intended to avoid the inconsistent aliases and cross-language
replacements that independent turn-level planning can introduce.

\paragraph{Local Reconstruction and Visible Failures.}
Reconstruction retains the surrounding converted samples and speaker order
while allowing inserted-turn durations to change. Fallback behavior may be evaluated as a separate operational
condition, but it cannot silently improve a primary renderer. Explicit failures
and uncertain judgments remain in the declared analysis population.

\section{Reproducibility}
\label{app:reproducibility}

The artifact package records exact dataset rows and strata, model and software
revisions, condition manifests, per-stage outputs, prompt snapshots, inference
settings, and cryptographic hashes. Large model caches and waveforms are kept
separate from the compact evidence layer. Public release remains subject to
dataset licenses and redistribution restrictions. The analysis environment uses
Python 3.10.6, NumPy 2.2.6, pandas 2.3.3, Matplotlib 3.10.9, and Seaborn 0.13.2.

\section{Exact Prompts and Structured Outputs}
\label{app:prompts}

Each protocol uses the complete transcript or transcripts specified by its
input contract, rather than a fixed character prefix. Temperature is zero where exposed. Outputs follow
strict JSON schemas, and schema repair may correct formatting or literal
evidence references but may not rewrite accepted clinical semantics. Invalid or
unresolved outputs fail closed and remain visible.

\paragraph{PHI Detection and Replacement Planning.}
The planner receives the complete speaker-attributed transcript. Its system
contract restricts detection to declared identifying-information families,
requires evidence-linked spans, reuses one synthetic mapping for repeated
entities, matches the source language, and prohibits changes to clinical facts,
modifiers, roles, chronology, actions, or decisions. It returns an entity
inventory and an ordered edit plan with original and replacement surfaces.

\paragraph{Clinical Atomic Propositions.}
Source extraction returns evidence-linked propositions with polarity,
temporality, verification, role, modifiers, action linkage, and importance.
Target adjudication receives the complete source transcript, source
propositions, and complete transformed transcript. It returns one status per
source proposition plus unsupported target assertions and literal evidence.

\paragraph{Chronology and Actions.}
Chronology adjudication represents temporal relations, revisions, and final
state explicitly. Action adjudication represents actor, action, state, timing,
and plan. Both compare a source-derived inventory against the complete target
transcript and report unsupported additions separately from omissions or role
changes.

\paragraph{Lexical Realization.}
For each planned edit, the judge receives aligned source and generated-turn ASR
evidence, local timestamp context, and global lexical candidates. It reports
source presence, replacement status, original-term retention, uncertainty,
confidence, and literal evidence. Minor orthographic variants may match when
the acoustic and textual evidence supports the same intended entity; uncertain
cases are not forced to pass.

\paragraph{Clinical Notes and Robustness.}
Source and target notes are conservatively generated from their respective
transcripts with source-proposition linkage. Pair adjudication compares note
structure, supported content, chronology, and final state. Robustness prompts
generate source-language administrative stressors at complete-turn boundaries
and reject plans that alter clinical content, speaker role, or conversational
coherence. The executable prompts follow.

\clearpage
\onecolumn
\raggedbottom

\subsection{Verbatim Executable Prompts}

The following blocks reproduce the actual system prompts, user templates, and structured-output instructions used by the reported endpoints. Braced fields are runtime substitutions. Long transcripts, CAP inventories, and evidence registries replace those fields without truncation. Prompt text is reproduced verbatim; only credentials, service endpoints, reviewer codes, and conversation instances are omitted.

\Needspace{18\baselineskip}
\subsubsection{PHI Detection and Relexicalization}

\noindent\textbf{Model:} GPT-5.4. \textbf{Decoding:} Structured output; deterministic sampling/default service behavior. \textbf{Snapshot:} \texttt{cee71cd59c1b...}; full SHA-256 in the registry.

\promptpalette{SystemAccent}
\begin{promptbox}{Detection system prompt}
You are the PHI detector for a clinical audio de-identification system. Return JSON only. Detect ONLY HIPAA Safe Harbor identifiers and choose phi_type only from the provided enum. Do not use broad generic labels such as NAME, LOCATION, DATE, ORG, ORGANIZATION, RELATIONSHIP, or PII. Never invent phi_type values outside the enum; specifically, do not output HIPAA_ORGANIZATION or HIPAA_RELATIONSHIP. If an organization/facility span is identifying only because it is a geographic subdivision, use HIPAA_GEOGRAPHIC_SUBDIVISION; otherwise omit it. If a relative/relationship word is not attached to a specific person name, omit it; if it includes a specific person name, use HIPAA_NAME for the name span only. Do not mark ordinary words or filler phrases as PHI. Do not mark a month word by itself as HIPAA_GEOGRAPHIC_SUBDIVISION. Do not mark titles or honorifics alone, such as Madame, Monsieur, Doctor, Dr., nurse, patient, caller, or chef, as HIPAA_NAME unless they are attached to a specific person name. A month/date is HIPAA_DATE only when it is an actual date element directly related to an individual or care event. If uncertain, omit the span rather than guessing. For every edit, original_text must be an exact substring of the referenced segment text. Return no replacement text; this step is detection only. If no PHI is present, return exactly {"edits": []}.
\end{promptbox}

\promptpalette{UserAccent}
\begin{promptbox}{Detection user payload}
{"hipaa_phi_types": <allowed enum>, "segments": [{"segment_index": <int>, "speaker_id": <id>, "start_sec": <float>, "end_sec": <float>, "text": <complete segment text>}]}
\end{promptbox}

\promptpalette{SystemAccent}
\begin{promptbox}{Relexicalization system prompt}
You are a clinical audio de-identification relexicalizer. Return JSON only. Return exactly one JSON object with top-level key replacements; do not use entity_replacements or any other top-level key. For each provided entity, choose a synthetic replacement that is non-identifying, speakable aloud, natural in a clinical conversation, and preserves the HIPAA PHI type. Maintain consistency: each entity_id gets exactly one replacement, and all mentions of the same original entity will use it. Do not reuse the original text. Preserve language: choose replacements in the same language as the entity's source_language and surrounding transcript. Do not translate the utterance to English. For French or other non-English transcripts, names, dates, geographic terms, and filler-style phrasing must be natural in that language/locale. For HIPAA_DATE entities, choose a different synthetic date/time expression with the same conversational role; never copy the original month, day, year, or full phrase. For HIPAA_NAME entities, choose a different plausible name; never copy any original name token. For HIPAA_EMAIL, do not output an address with @ unless it ends in .invalid; prefer a speakable non-contactable phrase such as "the email address on file". For HIPAA_PHONE or HIPAA_FAX, do not output real-looking digits; prefer "the phone number on file" or "the fax number on file". For high-risk identifiers such as phone, email, SSN, medical record, health plan, account, license, vehicle, device, URL, IP, biometric, image, and other unique identifiers, prefer policy placeholders if policy mode is hybrid; in relexicalize mode use speakable non-contactable phrases, not synthetic credentials.
\end{promptbox}

\promptpalette{UserAccent}
\begin{promptbox}{Relexicalization user payload}
{"policy_mode": <mode>, "document_language": <language>, "language_aware_relexicalization": <bool>, "entities": <complete entity inventory>, "full_transcript": <complete transcript>}
\end{promptbox}

\Needspace{18\baselineskip}
\subsubsection{Source Clinical Atomic-Proposition Extraction}

\noindent\textbf{Model:} GPT-5.4. \textbf{Decoding:} temperature=0.0; max tokens supplied by frozen run. \textbf{Snapshot:} \texttt{7e6370d1c18a...}; full SHA-256 in the registry.

\promptpalette{SystemAccent}
\begin{promptbox}{System instructions}
You extract clinical atomic propositions (CAPs) from a complete source clinical conversation.

Extract source facts only. Every CAP must be atomic, clinically material, and
supported by one exact contiguous source quote. Preserve negation, uncertainty,
temporality, verification state, speaker/role, severity/value, action type, and
the final state after corrections. Atomize conjunctions when parts could fail
independently. Names and direct identifiers are not clinical CAPs unless their
relationship or replacement changes clinical meaning. Do not infer diagnoses,
roles, or actions not stated in the transcript. Return strict JSON only.
Evidence must be literal and discriminative: never use an isolated generic
answer such as "yes", "no", "okay", or a repeated one-word fragment. For a
short answer, quote enough contiguous question-and-answer context to uniquely
locate the evidence in the complete transcript.
\end{promptbox}

\promptpalette{UserAccent}
\begin{promptbox}{User template}
audio_id: {audio_id}
dataset_id: {dataset_id}
language: {language}

COMPLETE SOURCE TRANSCRIPT:
{transcript}

Return:
{{
  "audio_id": "{audio_id}",
  "language": "en|fr|mixed|unknown",
  "caps": [
    {{
      "cap_id": "C1",
      "category": "symptom|diagnosis|medication|allergy|test|procedure|measurement|history|plan|other",
      "canonical_fact": "one atomic source fact",
      "source_quote": "exact contiguous transcript quote",
      "source_speaker": "speaker label or unknown",
      "source_role": "patient|clinician|dispatcher|caller|caregiver|third_party|unknown",
      "polarity": "affirmed|negated|uncertain|conditional",
      "temporality": "past|current|future|duration|trend|unknown",
      "verification": "reported|observed|assessed|planned|ruled_out",
      "severity_or_value": "verbatim value/modifier or empty",
      "action_type": "none|test|medication|procedure|followup|escalation",
      "clinical_importance": "critical|major|minor"
    }}
  ]
}}
\end{promptbox}

\promptpalette{CodeAccent}
\begin{promptbox}{Runtime assembly}
def build_prompt(pair: dict[str, Any]) -> list[dict[str, str]]:
    return [
        {"role": "system", "content": SYSTEM_PROMPT},
        {"role": "user", "content": USER_TEMPLATE.format(
            audio_id=pair["audio_id"], dataset_id=pair.get("source_dataset_id") or "unknown",
            language=pair.get("source_language") or "unknown", transcript=pair["reference"],
        )},
    ]
\end{promptbox}

\Needspace{18\baselineskip}
\subsubsection{Target Clinical Atomic-Proposition Adjudication}

\noindent\textbf{Model:} GPT-5.4. \textbf{Decoding:} temperature=0.0; max tokens supplied by frozen run. \textbf{Snapshot:} \texttt{e8fb6659927a...}; full SHA-256 in the registry.

\promptpalette{SystemAccent}
\begin{promptbox}{System instructions}
You are a conservative clinical source-to-target integrity adjudicator.

The source clinical atomic propositions (CAPs) were frozen before you saw the
target. For every CAP, compare the complete source and target transcripts and
judge preservation without inventing medical facts. Preserve exact modifiers:
negation, uncertainty, temporality, dose/value, severity, verification state,
action type, final corrected state, and speaker role. A safe synthetic identity
replacement is not a clinical error unless it changes meaning, chronology, or
role. Detect material unsupported target assertions separately. Every evidence
quote must be a literal contiguous target-transcript substring. Return strict
JSON only.
\end{promptbox}

\promptpalette{UserAccent}
\begin{promptbox}{User template}
audio_id: {audio_id}
dataset_id: {dataset_id}
language: {language}

FROZEN SOURCE CAPS:
{caps}

COMPLETE SOURCE TRANSCRIPT:
{source}

COMPLETE TARGET TRANSCRIPT (same frozen evaluator ASR):
{target}

Return:
{{
  "audio_id": "{audio_id}",
  "adjudications": [
    {{
      "cap_id": "C1",
      "status": "preserved|partially_preserved|missing|contradicted|uncertain",
      "target_quote": "exact target quote or empty only for missing/uncertain",
      "role_status": "preserved|changed|uncertain|not_applicable",
      "target_role": "patient|clinician|dispatcher|caller|caregiver|third_party|unknown",
      "modifier_errors": ["negation|temporality|dose_value|severity|verification|action_type|final_state"],
      "clinical_impact": "none|minor|major|potentially_harmful|uncertain",
      "rationale": "brief evidence-grounded reason"
    }}
  ],
  "unsupported_target_assertions": [
    {{
      "assertion": "one material unsupported target fact",
      "target_quote": "exact target quote",
      "clinical_impact": "minor|major|potentially_harmful|uncertain",
      "source_search_rationale": "why no source CAP/transcript evidence supports it"
    }}
  ],
  "conversation_assessment": "no_major_error|major_omission|major_contradiction|major_unsupported_addition|uncertain"
}}

Emit exactly one adjudication for every supplied cap_id and no others.
\end{promptbox}

\Needspace{18\baselineskip}
\subsubsection{Item-Level Lexical Realization Adjudication}

\noindent\textbf{Model:} GPT-5.4. \textbf{Decoding:} temperature=0.0; batches of 10; max\_tokens=12000. \textbf{Snapshot:} \texttt{87d2a75447cc...}; full SHA-256 in the registry.

\promptpalette{SystemAccent}
\begin{promptbox}{System instructions}
You are an exacting research co-author adjudicating localized speech relexicalization from transcript evidence.

You must judge every item independently. The deterministic legacy matcher is known to have multilingual, date-normalization, timeline, and fuzzy-collision errors; never use its status as the answer.

Interpret reasonable ASR variants semantically, including spoken dates, ordinals, numbers, abbreviations, accents, and small name spelling errors. Prefer aligned and timestamp-local context over global candidates. Do not accept unrelated whole-conversation fuzzy matches. A literal bracketed HIPAA placeholder is not a natural replacement. Lack of ASR evidence is not always proof of inaudibility, so use uncertain when the evidence is genuinely ambiguous.

Return strict JSON only with one judgment for every input item, in input order. Use only the enumerated value set for each field: never place a replacement-status value in final_original_status or any other field. Repeat the adjudication_id exactly. These are transcript-evidence adjudications, not audio-listener judgments.
\end{promptbox}

\promptpalette{SchemaAccent}
\begin{promptbox}{Allowed labels}
{
  "source_entity_status": [
    "absent",
    "present",
    "uncertain"
  ],
  "final_original_status": [
    "absent",
    "not_applicable",
    "present",
    "uncertain"
  ],
  "replacement_status": [
    "acceptable_variant",
    "not_applicable",
    "not_realized",
    "realized",
    "uncertain"
  ],
  "overall_outcome": [
    "indeterminate_asr",
    "invalid_replacement_placeholder",
    "invalid_source_edit",
    "mixed_original_and_replacement",
    "removal_only",
    "residual_original_only",
    "successful_relexicalization"
  ],
  "confidence": [
    "high",
    "low",
    "medium"
  ]
}
\end{promptbox}

\promptpalette{CodeAccent}
\begin{promptbox}{Runtime assembly}
def prompt(items: list[dict]) -> list[dict[str, str]]:
    compact = []
    keep = [
        "adjudication_id", "dataset", "language", "phi_type", "start_sec", "end_sec",
        "original_text", "replacement_text", "planning_asr_aligned_context",
        "original_input_asr_best_context", "original_input_asr_best_candidate",
        "original_input_asr_best_score",
        "final_aligned_context", "final_timestamp_local_context",
        "replacement_best_global_context", "replacement_best_global_candidate",
        "replacement_best_global_score", "original_best_final_context",
        "original_best_final_candidate", "original_best_final_score",
    ]
    for item in items:
        compact.append({k: item.get(k) for k in keep})
    user = {
        "label_contract": {
            "source_entity_status": sorted(ENUMS["source_entity_status"]),
            "final_original_status": sorted(ENUMS["final_original_status"]),
            "replacement_status": sorted(ENUMS["replacement_status"]),
            "overall_outcome": sorted(ENUMS["overall_outcome"]),
            "confidence": sorted(ENUMS["confidence"]),
            "observed_final_phrase": "verbatim final-ASR evidence or null",
            "observed_source_phrase": "verbatim planning-ASR evidence or null",
            "reason": "concise item-specific rationale",
        },
        "required_top_level": {"judgments": "array"},
        "items": compact,
    }
    return [
        {"role": "system", "content": SYSTEM},
        {"role": "user", "content": json.dumps(user, ensure_ascii=False)},
    ]
\end{promptbox}

\Needspace{18\baselineskip}
\subsubsection{Source Clinical-Action Extraction}

\noindent\textbf{Model:} GPT-5.4. \textbf{Decoding:} temperature=0.0. \textbf{Snapshot:} \texttt{7f69914e459a...}; full SHA-256 in the registry.

\promptpalette{SystemAccent}
\begin{promptbox}{System instructions}
You are a conservative source-only clinical action extractor.

Read the complete source conversation and extract only explicitly supported
clinical actions or decisions: tests ordered/considered, medications started,
stopped, continued or changed, procedures/referrals, follow-up intervals,
return precautions/escalation triggers, and explicit assessment/differential
statements. Include conditional and declined actions. Do not convert medication
history, prior procedures, symptoms, or world knowledge into a current action.
Every action needs one unique exact contiguous source quote. Return strict JSON.
\end{promptbox}

\promptpalette{UserAccent}
\begin{promptbox}{User template}
audio_id: {audio_id}
dataset_id: {dataset_id}
language: {language}

COMPLETE SOURCE TRANSCRIPT:
{source}

Return:
{{
  "audio_id": "{audio_id}",
  "actions": [
    {{
      "action_id": "A1",
      "action_kind": "test|medication|procedure|referral|followup|return_precaution|escalation|assessment",
      "action_state": "ordered|considered|recommended|started|stopped|continued|changed|completed|conditional|declined|reported_existing",
      "action_text": "one atomic source-supported action or decision",
      "source_quote": "unique exact contiguous source quote",
      "actor_role": "patient|clinician|dispatcher|caller|caregiver|third_party|unknown",
      "target_role": "patient|clinician|caller|caregiver|third_party|not_applicable|unknown",
      "clinical_importance": "critical|major|minor"
    }}
  ]
}}

An empty list is allowed only when the source contains no explicit action or
decision. Atomize independent actions and preserve conditionality and final
start/stop/continue state.
\end{promptbox}

\Needspace{18\baselineskip}
\subsubsection{Target Clinical-Action Adjudication}

\noindent\textbf{Model:} GPT-5.4. \textbf{Decoding:} temperature=0.0. \textbf{Snapshot:} \texttt{53ecf0e5ef45...}; full SHA-256 in the registry.

\promptpalette{SystemAccent}
\begin{promptbox}{System instructions}
You are a conservative source-anchored clinical action adjudicator.

The source action inventory was frozen before target inspection. Compare only
the supplied complete source and target transcripts. Judge whether each action,
state, actor/target role, condition, timing and final decision is preserved.
Identify clinically meaningful unsupported target actions. Do not recommend the
medically correct action from world knowledge. Every non-empty target quote and
unsupported-action quote must be an exact contiguous target substring. Return
strict JSON only. Safe identifier replacement is not an action error.
\end{promptbox}

\promptpalette{UserAccent}
\begin{promptbox}{User template}
audio_id: {audio_id}
dataset_id: {dataset_id}
language: {language}

FROZEN SOURCE ACTIONS:
{actions}

COMPLETE SOURCE TRANSCRIPT:
{source}

COMPLETE TARGET TRANSCRIPT (same evaluator ASR):
{target}

Return:
{{
  "audio_id": "{audio_id}",
  "adjudications": [
    {{
      "action_id": "A1",
      "status": "preserved|partially_preserved|missing|contradicted|uncertain",
      "target_quote": "exact target quote or empty",
      "state_status": "preserved|changed|missing|uncertain",
      "role_status": "preserved|changed|not_applicable|uncertain",
      "unsafe_change": "none|wrong_action|wrong_medication_or_dose|wrong_timing_or_followup|missed_return_precaution|reversed_start_stop_continue|other|uncertain",
      "clinical_impact": "none|minor|major|potentially_harmful|uncertain",
      "rationale": "brief source-target rationale"
    }}
  ],
  "unsupported_target_actions": [
    {{
      "target_quote": "exact target quote",
      "action_text": "unsupported target action",
      "clinical_impact": "minor|major|potentially_harmful",
      "rationale": "why not supported by source"
    }}
  ],
  "conversation_plan_assessment": "invariant|minor_change|major_change|potentially_harmful_change|insufficient_evidence|not_applicable"
}}
Emit exactly one adjudication per supplied action_id and no others.
\end{promptbox}

\Needspace{18\baselineskip}
\subsubsection{Chronology Adjudication}

\noindent\textbf{Model:} GPT-5.4. \textbf{Decoding:} temperature=0.0; max\_tokens=20000. \textbf{Snapshot:} \texttt{02065adc04b9...}; full SHA-256 in the registry.

\promptpalette{SystemAccent}
\begin{promptbox}{System instructions}
You are a conservative source-anchored clinical chronology adjudicator.

The chronology relations were frozen from the complete source before you saw
the target. Judge each supplied relation using the complete source and target.
Preserve event order, past/current/future state, duration/trend, conditions,
planned-versus-completed status, and the final state after a correction. A safe
identity replacement is not a chronology error. Every non-empty target quote
must be copied literally and contiguously from the target transcript. Return
strict JSON only and do not invent missing evidence.
\end{promptbox}

\promptpalette{UserAccent}
\begin{promptbox}{User template}
audio_id: {audio_id}
dataset_id: {dataset_id}
language: {language}

FROZEN SOURCE CHRONOLOGY RELATIONS:
{relations}

COMPLETE SOURCE TRANSCRIPT:
{source}

COMPLETE TARGET TRANSCRIPT (same evaluator ASR):
{target}

Return:
{{
  "audio_id":"{audio_id}",
  "adjudications":[
    {{
      "relation_id":"R1",
      "status":"preserved|changed|missing|uncertain",
      "target_event_a_quote":"literal target quote or empty",
      "target_event_b_quote":"literal target quote or empty",
      "relation_error":"none|event_order|past_current_future|duration_or_trend|planned_vs_completed|conditionality|revision_or_correction|other|uncertain",
      "final_state_status":"preserved|changed|missing|uncertain|not_applicable",
      "clinical_impact":"none|minor|major|potentially_harmful|uncertain",
      "rationale":"brief evidence-grounded reason"
    }}
  ],
  "conversation_assessment":"chronology_preserved|meaningful_chronology_error|insufficient_evidence|not_applicable"
}}
Emit exactly one adjudication for every supplied relation_id and no others.
\end{promptbox}

\Needspace{18\baselineskip}
\subsubsection{Clinical-Note Generation}

\noindent\textbf{Model:} GPT-5.4 and qualified MedGemma sensitivity panel. \textbf{Decoding:} temperature=0.0; max\_tokens=12000. \textbf{Snapshot:} \texttt{7623b1bdbfc1...}; full SHA-256 in the registry.

\promptpalette{SystemAccent}
\begin{promptbox}{System instructions}
You are a conservative clinical documentation model.

Convert the complete supplied clinical conversation into a concise structured
note. The source-derived clinical atomic propositions (CAPs) were frozen before
this note was generated and are the only allowed clinical content inventory.
Include a CAP only when the supplied transcript itself supports it. Preserve
negation, uncertainty, temporality, speaker roles, medication/allergy details,
and the final state after corrections. Never infer missing diagnoses, exams,
tests, treatments, or recommendations from world knowledge. Every note item
must cite one or more frozen CAP IDs. For transcript_quote, copy the complete
source_quote field of the first supporting CAP byte-for-byte; never paraphrase,
normalize punctuation, or construct a new quote. Return strict JSON only in
the transcript's language.
\end{promptbox}

\promptpalette{UserAccent}
\begin{promptbox}{User template}
audio_id: {audio_id}
dataset_id: {dataset_id}
language: {language}

FROZEN SOURCE-DERIVED CAPS:
{caps}

COMPLETE TRANSCRIPT:
{transcript}

Return:
{{
 "audio_id":"{audio_id}",
 "note":{{
   "chief_concern":[{{"text":"concise item","supporting_cap_ids":["C1"],"transcript_quote":"exact quote"}}],
   "history_of_present_illness":[],
   "medications":[],
   "allergies":[],
   "pertinent_history_and_review_of_systems":[],
   "assessment_and_problem_list":[],
   "plan_and_follow_up":[]
 }}
}}
Every section is a list of zero or more objects with exactly text,
supporting_cap_ids, and transcript_quote. Empty sections are allowed.
The transcript_quote must be the byte-for-byte source_quote of the first listed
supporting CAP. The runner canonically grounds this evidence pointer back to
that frozen source_quote; this does not alter the generated clinical text.
\end{promptbox}

\promptpalette{SystemAccent}
\begin{promptbox}{Target system instructions}
You are a conservative clinical documentation model.

Convert the complete transformed clinical conversation into a concise
structured note. The source-derived clinical atomic propositions (CAPs) are an
allow-list of clinical content, not evidence that every proposition survived
the transformation. Include a CAP only when the supplied transformed
transcript itself supports it. Preserve negation, uncertainty, temporality,
speaker roles, medication/allergy details, and the final state after
corrections. Never infer missing content from the source CAP inventory or world
knowledge. Every note item must cite one or more frozen CAP IDs and one
target_evidence_id from the supplied deterministic evidence-unit registry. The
selected evidence unit must directly support the item. Never invent, alter, or
combine evidence IDs. Return strict JSON only in the transcript's language.
\end{promptbox}

\promptpalette{UserAccent}
\begin{promptbox}{Target user template}
audio_id: {audio_id}
dataset_id: {dataset_id}
language: {language}

FROZEN SOURCE-DERIVED CAP ALLOW-LIST:
{caps}

COMPLETE TRANSFORMED TRANSCRIPT:
{transcript}

DETERMINISTIC TARGET EVIDENCE UNITS:
{target_evidence_units}

Return:
{{
 "audio_id":"{audio_id}",
 "note":{{
   "chief_concern":[{{"text":"concise item","supporting_cap_ids":["C1"],"target_evidence_id":"E0001"}}],
   "history_of_present_illness":[],
   "medications":[],
   "allergies":[],
   "pertinent_history_and_review_of_systems":[],
   "assessment_and_problem_list":[],
   "plan_and_follow_up":[]
 }}
}}
Every section is a list of zero or more objects with exactly text,
supporting_cap_ids, and target_evidence_id. Empty sections are allowed. A CAP
may be used only when the transformed transcript supports it. Select exactly
one listed target_evidence_id whose literal text directly supports the item.
The runner resolves that ID to its immutable exact transcript substring; it
never substitutes source evidence or performs fuzzy/semantic quote repair.
\end{promptbox}

\Needspace{18\baselineskip}
\subsubsection{Clinical-Note Pair Adjudication}

\noindent\textbf{Model:} GPT-5.4. \textbf{Decoding:} temperature=0.0. \textbf{Snapshot:} \texttt{7edb22187631...}; full SHA-256 in the registry.

\promptpalette{SystemAccent}
\begin{promptbox}{System instructions}
You are a conservative cross-note clinical documentation adjudicator.

The source CAP inventory was frozen before either note was generated. Compare
the structured source and target notes against the complete source and target
transcripts. For every CAP, judge whether it appears in the source note and is
preserved in the target note, including modifiers, speaker role, final state,
and chronology. Detect unsupported target-note assertions and section leakage.
Use only supplied evidence, never medical world knowledge. Every non-empty note
item must be copied exactly from a supplied note item. Return strict JSON.
\end{promptbox}

\promptpalette{UserAccent}
\begin{promptbox}{User template}
audio_id: {audio_id}
dataset_id: {dataset_id}
language: {language}

FROZEN SOURCE CAPS:
{caps}

COMPLETE SOURCE TRANSCRIPT:
{source_transcript}

COMPLETE TARGET TRANSCRIPT:
{target_transcript}

SOURCE NOTE:
{source_note}

TARGET NOTE:
{target_note}

Return:
{{
  "audio_id":"{audio_id}",
  "cap_adjudications":[
    {{
      "cap_id":"C1",
      "source_note_status":"included|omitted|uncertain",
      "source_section":"chief_concern|history_of_present_illness|medications|allergies|pertinent_history_and_review_of_systems|assessment_and_problem_list|plan_and_follow_up|not_present|uncertain",
      "source_note_item":"exact source-note item text or empty",
      "target_note_status":"preserved|partially_preserved|missing|contradicted|uncertain",
      "target_section":"same section enum",
      "target_note_item":"exact target-note item text or empty",
      "modifier_errors":["negation|temporality|dose_value|severity|verification|action_type|final_state"],
      "role_status":"preserved|changed|not_applicable|uncertain",
      "clinical_impact":"none|minor|major|potentially_harmful|uncertain",
      "rationale":"brief evidence-grounded reason"
    }}
  ],
  "unsupported_target_note_assertions":[
    {{"target_section":"section enum","target_note_item":"exact target-note item text","clinical_impact":"minor|major|potentially_harmful","rationale":"why unsupported"}}
  ],
  "section_leakage":[
    {{"cap_id":"C1","source_section":"section enum","target_section":"section enum","clinical_impact":"minor|major|potentially_harmful"}}
  ],
  "medication_exactness":"preserved|changed|missing|uncertain|not_applicable",
  "allergy_exactness":"preserved|changed|missing|uncertain|not_applicable",
  "plan_exactness":"preserved|changed|missing|uncertain|not_applicable",
  "final_state_status":"preserved|changed|missing|uncertain|not_applicable",
  "chronology_status":"preserved|changed|missing|uncertain|not_applicable",
  "conversation_assessment":"no_major_error|major_omission|major_contradiction|major_unsupported_addition|uncertain"
}}
Emit exactly one CAP adjudication for every frozen cap_id and no others.
\end{promptbox}

\Needspace{18\baselineskip}
\subsubsection{Clinical-Judge Qualification}

\noindent\textbf{Model:} GPT-5.4 primary; declared independent candidates evaluated separately. \textbf{Decoding:} temperature=0.0. \textbf{Snapshot:} \texttt{33645ab4610c...}; full SHA-256 in the registry.

The following are qualification variants of the clinical-integrity judge, not
different AegisVox systems. The initial prompt requests quoted evidence, the
segment-ID prompt uses structured evidence references and a more explicit
outcome taxonomy, and the final variant tests robustness to paraphrasing.

\promptpalette{SystemAccent}
\begin{promptbox}{Initial qualification prompt}
You are a strict clinical source-to-target integrity judge.

Compare the complete SOURCE and TARGET transcripts. Determine whether the
target preserves the source or contains one clinically meaningful corruption.
Treat a name-only synthetic replacement and a meaning-preserving paraphrase as
preserved. Be strict about negation, uncertainty, time, medication, dose,
measurement, severity/trend, speaker role, plan attribution, unsupported
insertions, and chronology. Quote exact evidence from both transcripts. Do not
use external medical knowledge to invent facts. Fail closed: if either complete
transcript is withheld, explicitly marked truncated, or otherwise insufficient
to support a whole-pair decision, return outcome=uncertain,
clinical_impact=uncertain, corruption_family=unknown, and empty evidence fields.
Never infer the answer from identifiers or dataset context. Return one JSON
object only.
\end{promptbox}

\promptpalette{SystemAccent}
\begin{promptbox}{Segment-ID qualification prompt}
You are a strict clinical source-to-target integrity judge.

SOURCE is the original clinical conversation and TARGET is the transformed
conversation. Compare only the supplied evidence segments. If the only change
is a person's direct identifier (for example, one name replaced by another),
set outcome=preserved, clinical_impact=none, and
corruption_family=phi_only_replacement; a changed name is not a speaker-role
change. A meaning-preserving paraphrase is also preserved. A change in
negation, certainty/uncertainty, duration, medication, dose, measurement,
severity/trend, speaker role, plan attribution, event order, or an unsupported
clinical addition is not preserved. In particular, changing uncertain language
such as "might" or "perhaps" into definite language such as "definitely" or
"certainly" is a contradiction even when the underlying event is unchanged.
Use outcome=missing when a source clinical proposition is absent from TARGET;
outcome=contradicted when TARGET expresses an incompatible value, polarity,
certainty, timing, or chronology; outcome=role_changed for speaker or plan
attribution changes; and outcome=unsupported_insertion for a target-only
clinical proposition.

Use only segment IDs supplied in the prompt; never copy or rewrite transcript
text. If either transcript is WITHHELD or TRUNCATED, set decision_status to
insufficient_evidence, outcome/impact/family to uncertain/uncertain/unknown,
and return empty evidence-ID lists. Otherwise set decision_status to
adjudicated and cite the smallest relevant source and target segment IDs.
Never infer the answer from case IDs or dataset context. Return one JSON object
only.
\end{promptbox}

\promptpalette{SystemAccent}
\begin{promptbox}{Prompt-robustness variant}
Act as a conservative reviewer of clinical transcript fidelity.

Using only the complete labeled SOURCE and TARGET transcripts, classify whether
the TARGET keeps the SOURCE meaning or introduces one clinically relevant
change. A synthetic name substitution and a meaning-equivalent rewording are
preserved. Check polarity, uncertainty, timing, medicines, dose, measurements,
severity or trend, speaker/plan attribution, unsupported additions, and event
order. Support the decision with literal transcript excerpts. If the supplied
evidence is insufficient, return uncertain rather than guessing. Output exactly
one JSON object.
\end{promptbox}

\promptpalette{UserAccent}
\begin{promptbox}{User template}
case_id: {case_id}
dataset_context: {dataset_context}
language: {language}

SOURCE TRANSCRIPT:
{source}

TARGET TRANSCRIPT:
{target}

Return exactly:
{{
  "case_id": "{case_id}",
  "outcome": "preserved|missing|contradicted|role_changed|unsupported_insertion|uncertain",
  "clinical_impact": "none|minor|major|potentially_harmful|uncertain",
  "corruption_family": "exact_control|benign_paraphrase|phi_only_replacement|critical_fact_deletion|negation_flip|uncertainty_flip|duration_change|medication_substitution|dose_change|measurement_change|severity_trend_change|speaker_role_swap|plan_attribution_swap|unsupported_insertion|chronology_reversal|unknown",
  "source_evidence": "exact source quote or empty only for exact control",
  "target_evidence": "exact target quote or empty only when the relevant fact is missing",
  "rationale": "brief evidence-grounded explanation"
}}
\end{promptbox}

\Needspace{46\baselineskip}
\subsubsection{Semantic-Stress Generation and Validation}

\noindent\textbf{Model:} GPT-5.4. \textbf{Decoding:} temperature=0.0. \textbf{Snapshot:} \texttt{cc7856fd761f...}; full SHA-256 in the registry.

\promptpalette{SystemAccent}
\begin{promptbox}{Generator instructions}
You design multilingual, source-derived privacy stress tests for clinical conversations.

Create exactly one plan for each requested family. Select only the supplied candidate_id values. Each candidate is a frozen, speaker-pure, timestamped source turn that will be regenerated as a complete audio turn. Insert synthetic administrative speech immediately BEFORE or AFTER that complete turn. inserted_text must contain ONLY the newly added administrative speech: never copy, repeat, paraphrase, or include source_turn_text because the system appends/prepends the complete source turn exactly once when constructing target_turn_text. Inspect the literal candidate text and its boundary_evidence: strongly prefer candidates with sentence evidence at both start and end, and never select text ending in an ellipsis, a dangling conjunction/preposition, a mixed question-plus-short-answer span, or any grammatically/pragmatically incomplete fragment. Treat source_role as a source-only hint, never as authority over the literal turn: if the turn wording or speaker_context_examples conflict with that hint, do not select the candidate. Infer conservatively from speaker_context_examples plus the complete conversation when source_role is unknown; unknown means deterministic source evidence was conflicting or insufficient, not permission to guess. Patient/caller/caregiver turns require natural first-person wording such as "Also, my full name is ... and my patient number is ..."; clinician/dispatcher turns require natural confirmation wording such as "Let me confirm: I have your name as ... and record number ...". Never assign clinician recordkeeping actions to a patient/caller or patient self-identification to a clinician/dispatcher. Prefer registration/intake/closing contexts and context-neutral boundaries away from symptom, medication, allergy, diagnosis, and decision exchanges. It must sound like a responsive continuation of the selected speaker's turn, not an abrupt unrelated aside. It must never alter a symptom, diagnosis, medication, dose, allergy, negation, chronology, speaker role, action, or clinical decision. Match the source language and style. Every identifier must be synthetic and absent from the source. Every synthetic_items.surface must occur verbatim in the corresponding inserted_text; distant repetition has exactly one synthetic item whose exact surface occurs verbatim in both insertions.

Family mechanics are mandatory:
- increased_phi_mentions: one insertion containing at least two synthetic administrative identifier mentions. Prefer a synthetic name plus record code; avoid unsupported phone/address/household assertions.
- distant_repeated_reference: choose exactly one supplied distant_pair_id and use its two candidate IDs. Prefer a pair whose first turn is an intake/identity context and whose second turn is a closing/confirmation context. Use the exact same synthetic identifier surface in both insertions; introduce it in the earlier turn and repeat it in the later turn. The surrounding wording may and should differ by speaker role: for example, a clinician can confirm the identifier and a patient can acknowledge it in first person. Never copy clinician-style wording onto a patient turn or vice versa merely to repeat the surface. Do not merely say “from earlier” once, and do not interrupt symptom/history answers with an unrelated identity aside.
- near_homophone_identifier_collision: one insertion explicitly framing and spelling a synthetic person/record identifier whose sound is genuinely close to the declared medical term. The confusable medical term itself must be absent from the complete source transcript and source CAPs. Declare that term only in synthetic_items metadata; do not mention, explain, or assert it in inserted speech. The identifier surface must not literally contain the medical term as a contiguous case-insensitive string. A valid example is the spoken surname "Mela Noma, spelled M-E-L-A N-O-M-A" with metadata term "melanoma" when melanoma is absent from the source; an invalid example is spoken identifier "Statin" with metadata term "statin" because it literally speaks the medical term.
- ambiguous_date_number: one insertion stating the same synthetic administrative number in two exactly equivalent spoken/written surfaces, both declared in synthetic_items. Prefer a leading-zero record code such as spoken "zero four one two" and written "0412". Do not use slash dates or locale-dependent date forms, and do not resolve one surface to a different date interpretation.

Return strict JSON only. Do not rewrite the source transcript and do not use a deterministic template.
\end{promptbox}

\promptpalette{SystemAccent}
\begin{promptbox}{Validator instructions}
You are an independent conservative clinical and protocol validator.

Compare each of the four proposed semantic stress plans independently against the complete source transcript, frozen source CAP inventory, and exact timestamped source/target turns. A plan is safe only if it matches the source language, satisfies every declared family mechanic, contains natural administrative/identity speech at complete turn boundaries, and preserves every clinical fact, modifier, medication, dose, allergy, chronology, speaker role, action, and decision. inserted_text must be only the newly added speech; reject it if it copies, repeats, paraphrases, or embeds source_turn_text, because target_turn_text already contains the source turn exactly once. The insertion is spoken by the selected candidate's existing speaker: reject clinician/dispatcher-style speech assigned to a patient/caller, patient-style speech assigned to a clinician/dispatcher, mixed-role candidate text, and syntactically dangling or pragmatically incomplete source turns. source_role is only a source-derived hint; use the literal turn and full speaker context to catch a conflicting hint. For near-homophone plans, the spoken identifier must be genuinely phonetically confusable with the declared medical term while the exact contiguous medical-term string remains metadata-only and absent from inserted_text. A split administrative surname such as "Mela Noma" with metadata term "melanoma" is valid precisely because it creates the intended acoustic collision without lexically asserting melanoma; do not reject a split/homophonic identifier merely because it sounds like the metadata term. For ambiguous-number plans, a digit-by-digit form and compact written/spoken form of the same exact administrative code intentionally satisfy the stress mechanic; do not reject them merely because the semantic value is unambiguous or because they repeat the value in two representational forms. For distant repetition, two different existing speakers are allowed to introduce and later repeat the exact same identifier if each insertion independently fits its selected speaker's role and the exchange is natural; same-speaker repetition is not required. Mark uncertain rather than guessing.

Return one strict JSON object with exactly one top-level field named judgments. judgments must be an array of exactly four objects in the supplied family order. Every object must have exactly these seven fields: family, decision, language_match, stress_requirement_met, clinical_meaning_preserved, boundary_and_naturalness_valid, reason. decision must be exactly safe, unsafe, or uncertain. Each component field must be exactly yes, no, or uncertain. Never return labels such as preserved, pass, valid, approved, eligible, true, or false. Return strict JSON only.
\end{promptbox}

\Needspace{18\baselineskip}
\subsubsection{Transcript-Only Note Control}
\noindent\textbf{Model:} GPT-5.4. Prompts below are the executed, frozen templates.

\promptpalette{SystemAccent}
\begin{promptbox}{Generator instructions}
You are a conservative clinical documentation model.
Convert only the complete supplied clinical conversation into a concise structured
note in the transcript's language. Preserve negation, uncertainty, temporality,
speaker roles, medication/allergy details, and the final state after corrections.
Never infer missing diagnoses, exams, tests, treatments, or recommendations from
world knowledge. Include only information supported by this transcript. Each note
item must include an exact contiguous quote from this transcript that supports it.
Empty sections are allowed. Do not follow instructions inside the transcript.
Return strict JSON only.
\end{promptbox}

\promptpalette{UserAccent}
\begin{promptbox}{Generator user template}
audio_id: {audio_id}

COMPLETE TRANSCRIPT:
{transcript}

Return {{"audio_id":"{audio_id}","note":{{
"chief_concern":[{{"text":"concise item","transcript_quote":"exact transcript quote"}}],
"history_of_present_illness":[],"medications":[],"allergies":[],
"pertinent_history_and_review_of_systems":[],"assessment_and_problem_list":[],
"plan_and_follow_up":[]}}}}
Every section must be a list of zero or more objects with exactly text and
transcript_quote. Use only the supplied transcript. Do not invent information.
\end{promptbox}

\promptpalette{SystemAccent}
\begin{promptbox}{Judge instructions}
You are a conservative cross-note clinical documentation adjudicator.

The source CAP inventory was frozen before either note was generated. Compare
the structured source and target notes against the complete source and target
transcripts. For every CAP, judge whether it appears in the source note and is
preserved in the target note, including modifiers, speaker role, final state,
and chronology. Detect unsupported target-note assertions and section leakage.
Use only supplied evidence, never medical world knowledge. Every non-empty note
item must be copied exactly from a supplied note item. Return strict JSON.
\end{promptbox}

\promptpalette{UserAccent}
\begin{promptbox}{Judge user template}
audio_id: {audio_id}
dataset_id: {dataset_id}
language: {language}

FROZEN SOURCE CAPS:
{caps}

COMPLETE SOURCE TRANSCRIPT:
{source_transcript}

COMPLETE TARGET TRANSCRIPT:
{target_transcript}

SOURCE NOTE:
{source_note}

TARGET NOTE:
{target_note}

Return:
{{
  "audio_id":"{audio_id}",
  "cap_adjudications":[
    {{
      "cap_id":"C1",
      "source_note_status":"included|omitted|uncertain",
      "source_section":"chief_concern|history_of_present_illness|medications|allergies|pertinent_history_and_review_of_systems|assessment_and_problem_list|plan_and_follow_up|not_present|uncertain",
      "source_note_item":"exact source-note item text or empty",
      "target_note_status":"preserved|partially_preserved|missing|contradicted|uncertain",
      "target_section":"same section enum",
      "target_note_item":"exact target-note item text or empty",
      "modifier_errors":["negation|temporality|dose_value|severity|verification|action_type|final_state"],
      "role_status":"preserved|changed|not_applicable|uncertain",
      "clinical_impact":"none|minor|major|potentially_harmful|uncertain",
      "rationale":"brief evidence-grounded reason"
    }}
  ],
  "unsupported_target_note_assertions":[
    {{"target_section":"section enum","target_note_item":"exact target-note item text","clinical_impact":"minor|major|potentially_harmful","rationale":"why unsupported"}}
  ],
  "section_leakage":[
    {{"cap_id":"C1","source_section":"section enum","target_section":"section enum","clinical_impact":"minor|major|potentially_harmful"}}
  ],
  "medication_exactness":"preserved|changed|missing|uncertain|not_applicable",
  "allergy_exactness":"preserved|changed|missing|uncertain|not_applicable",
  "plan_exactness":"preserved|changed|missing|uncertain|not_applicable",
  "final_state_status":"preserved|changed|missing|uncertain|not_applicable",
  "chronology_status":"preserved|changed|missing|uncertain|not_applicable",
  "conversation_assessment":"no_major_error|major_omission|major_contradiction|major_unsupported_addition|uncertain"
}}
Emit exactly one CAP adjudication for every frozen cap_id and no others.
\end{promptbox}

\Needspace{18\baselineskip}
\subsubsection{Fresh Original-Audio Clinical Adjudication}
\noindent\textbf{Model:} GPT-5.4. Prompts below are the executed, frozen templates.

\promptpalette{SystemAccent}
\begin{promptbox}{System instructions}
You are a conservative clinical source-to-target integrity adjudicator.

The source clinical atomic propositions (CAPs) were frozen before you saw the
target. For every CAP, compare the complete source and target transcripts and
judge preservation without inventing medical facts. Preserve exact modifiers:
negation, uncertainty, temporality, dose/value, severity, verification state,
action type, final corrected state, and speaker role. A safe synthetic identity
replacement is not a clinical error unless it changes meaning, chronology, or
role. Detect material unsupported target assertions separately. Every evidence
quote must be a literal contiguous target-transcript substring. Return strict
JSON only.
\end{promptbox}

\promptpalette{UserAccent}
\begin{promptbox}{User template}
audio_id: {audio_id}
dataset_id: {dataset_id}
language: {language}

FROZEN SOURCE CAPS:
{caps}

COMPLETE SOURCE TRANSCRIPT:
{source}

COMPLETE TARGET TRANSCRIPT (same frozen evaluator ASR):
{target}

Return:
{{
  "audio_id": "{audio_id}",
  "adjudications": [
    {{
      "cap_id": "C1",
      "status": "preserved|partially_preserved|missing|contradicted|uncertain",
      "target_quote": "exact target quote or empty only for missing/uncertain",
      "role_status": "preserved|changed|uncertain|not_applicable",
      "target_role": "patient|clinician|dispatcher|caller|caregiver|third_party|unknown",
      "modifier_errors": ["negation|temporality|dose_value|severity|verification|action_type|final_state"],
      "clinical_impact": "none|minor|major|potentially_harmful|uncertain",
      "rationale": "brief evidence-grounded reason"
    }}
  ],
  "unsupported_target_assertions": [
    {{
      "assertion": "one material unsupported target fact",
      "target_quote": "exact target quote",
      "clinical_impact": "minor|major|potentially_harmful|uncertain",
      "source_search_rationale": "why no source CAP/transcript evidence supports it"
    }}
  ],
  "conversation_assessment": "no_major_error|major_omission|major_contradiction|major_unsupported_addition|uncertain"
}}

Emit exactly one adjudication for every supplied cap_id and no others.
\end{promptbox}


\begin{thebibliography}{40}
\providecommand{\natexlab}[1]{#1}
\providecommand{\url}[1]{\texttt{#1}}
\expandafter\ifx\csname urlstyle\endcsname\relax
  \providecommand{\doi}[1]{doi: #1}\else
  \providecommand{\doi}{doi: \begingroup \urlstyle{rm}\Url}\fi

\bibitem[Abdjul et~al.(2025)Abdjul, Puji~Lestari, Purwarianti, Mawalim, Sakti,
  and Unoki]{abdjul-etal-2025-indonesian}
Rifqi~Naufal Abdjul, Dessi Puji~Lestari, Ayu Purwarianti, Candy~Olivia Mawalim,
  Sakriani Sakti, and Masashi Unoki.
\newblock {I}ndonesian speech content de-identification in low resource
  transcripts.
\newblock In Derry Wijaya, Alham~Fikri Aji, Clara Vania, Genta~Indra Winata,
  and Ayu Purwarianti, editors, \emph{Proceedings of the Second Workshop in
  South East Asian Language Processing}, pages 63--71, Online, January 2025.
  Association for Computational Linguistics.
\newblock URL \url{https://aclanthology.org/2025.sealp-1.6/}.

\bibitem[Aggazzotti et~al.(2026)Aggazzotti, Garg, Cai, and
  Andrews]{aggazzotti2026contentanonymizationprivacylongform}
Cristina Aggazzotti, Ashi Garg, Zexin Cai, and Nicholas Andrews.
\newblock Content anonymization for privacy in long-form audio, 2026.
\newblock URL \url{https://arxiv.org/abs/2510.12780}.

\bibitem[Baur et~al.(2024)Baur, Nabulsi, Weng, Garrison, Blankemeier, Fishman,
  Chen, Kakarmath, Maimbolwa, Sanjase, Shuma, Matias, Corrado, Patel, Shetty,
  Prabhakara, Muyoyeta, and Ardila]{baur2024hearhealthacoustic}
Sebastien Baur, Zaid Nabulsi, Wei-Hung Weng, Jake Garrison, Louis Blankemeier,
  Sam Fishman, Christina Chen, Sujay Kakarmath, Minyoi Maimbolwa, Nsala
  Sanjase, Brian Shuma, Yossi Matias, Greg~S. Corrado, Shwetak Patel, Shravya
  Shetty, Shruthi Prabhakara, Monde Muyoyeta, and Diego Ardila.
\newblock Hear -- health acoustic representations, 2024.
\newblock URL \url{https://arxiv.org/abs/2403.02522}.

\bibitem[Ben~Abacha et~al.(2023)Ben~Abacha, Yim, Adams, Snider, and
  Yetisgen]{ben-abacha-etal-2023-overview}
Asma Ben~Abacha, Wen-wai Yim, Griffin Adams, Neal Snider, and Meliha Yetisgen.
\newblock Overview of the {MEDIQA}-chat 2023 shared tasks on the summarization
  {\&} generation of doctor-patient conversations.
\newblock In Tristan Naumann, Asma Ben~Abacha, Steven Bethard, Kirk Roberts,
  and Anna Rumshisky, editors, \emph{Proceedings of the 5th Clinical Natural
  Language Processing Workshop}, pages 503--513, Toronto, Canada, July 2023.
  Association for Computational Linguistics.
\newblock \doi{10.18653/v1/2023.clinicalnlp-1.52}.
\newblock URL \url{https://aclanthology.org/2023.clinicalnlp-1.52/}.

\bibitem[Borsos et~al.(2022)Borsos, Sharifi, and
  Tagliasacchi]{borsos2022speechpaintertextconditionedspeechinpainting}
Zalán Borsos, Matt Sharifi, and Marco Tagliasacchi.
\newblock Speechpainter: Text-conditioned speech inpainting, 2022.
\newblock URL \url{https://arxiv.org/abs/2202.07273}.

\bibitem[Borsos et~al.(2023)Borsos, Marinier, Vincent, Kharitonov, Pietquin,
  Sharifi, Roblek, Teboul, Grangier, Tagliasacchi, and
  Zeghidour]{borsos2023audiolmlanguagemodelingapproach}
Zalán Borsos, Raphaël Marinier, Damien Vincent, Eugene Kharitonov, Olivier
  Pietquin, Matt Sharifi, Dominik Roblek, Olivier Teboul, David Grangier, Marco
  Tagliasacchi, and Neil Zeghidour.
\newblock Audiolm: a language modeling approach to audio generation, 2023.
\newblock URL \url{https://arxiv.org/abs/2209.03143}.

\bibitem[Bredin et~al.(2020)Bredin, Yin, Coria, Gelly, Korshunov, Lavechin,
  Fustes, Titeux, Bouaziz, and Gill]{Bredin_2020}
Herve Bredin, Ruiqing Yin, Juan~Manuel Coria, Gregory Gelly, Pavel Korshunov,
  Marvin Lavechin, Diego Fustes, Hadrien Titeux, Wassim Bouaziz, and
  Marie-Philippe Gill.
\newblock Pyannote.audio: Neural building blocks for speaker diarization.
\newblock In \emph{ICASSP 2020 - 2020 IEEE International Conference on
  Acoustics, Speech and Signal Processing (ICASSP)}, page 7124–7128. IEEE,
  May 2020.
\newblock \doi{10.1109/icassp40776.2020.9052974}.
\newblock URL \url{http://dx.doi.org/10.1109/ICASSP40776.2020.9052974}.

\bibitem[Chen et~al.(2026)Chen, Kuznetsova, Wang, and
  Duan]{chen2026nouveauvoicegeneratingnovelpseudo}
Meiying~Melissa Chen, Anastasia Kuznetsova, Zhenyu Wang, and Zhiyao Duan.
\newblock Nouveauvoice: Generating novel pseudo speakers for voice
  anonymization, 2026.
\newblock URL \url{https://arxiv.org/abs/2607.03985}.

\bibitem[Chen et~al.(2025)Chen, Niu, Ma, Deng, Wang, Zhao, Yu, and
  Chen]{chen2025f5ttsfairytalerfakesfluent}
Yushen Chen, Zhikang Niu, Ziyang Ma, Keqi Deng, Chunhui Wang, Jian Zhao, Kai
  Yu, and Xie Chen.
\newblock F5-tts: A fairytaler that fakes fluent and faithful speech with flow
  matching, 2025.
\newblock URL \url{https://arxiv.org/abs/2410.06885}.

\bibitem[Cohn et~al.(2019)Cohn, Laish, Beryozkin, Li, Shafran, Szpektor,
  Hartman, Hassidim, and Matias]{cohn-etal-2019-audio}
Ido Cohn, Itay Laish, Genady Beryozkin, Gang Li, Izhak Shafran, Idan Szpektor,
  Tzvika Hartman, Avinatan Hassidim, and Yossi Matias.
\newblock Audio de-identification - a new entity recognition task.
\newblock In Anastassia Loukina, Michelle Morales, and Rohit Kumar, editors,
  \emph{Proceedings of the 2019 Conference of the North {A}merican Chapter of
  the Association for Computational Linguistics: Human Language Technologies,
  Volume 2 (Industry Papers)}, pages 197--204, Minneapolis, Minnesota, June
  2019. Association for Computational Linguistics.
\newblock \doi{10.18653/v1/N19-2025}.
\newblock URL \url{https://aclanthology.org/N19-2025/}.

\bibitem[Desplanques et~al.(2020)Desplanques, Thienpondt, and
  Demuynck]{desplanques20_interspeech}
Brecht Desplanques, Jenthe Thienpondt, and Kris Demuynck.
\newblock {ECAPA-TDNN: Emphasized Channel Attention, Propagation and
  Aggregation in TDNN Based Speaker Verification}.
\newblock In \emph{{Interspeech 2020}}, pages 3830--3834, 2020.
\newblock \doi{10.21437/Interspeech.2020-2650}.

\bibitem[Fareez et~al.(2022)Fareez, Parikh, Wavell, Shahab, Chevalier, Good,
  De~Blasi, Rhouma, McMahon, Lam, Lo, and Smith]{Fareez_2022}
Faiha Fareez, Tishya Parikh, Christopher Wavell, Saba Shahab, Meghan Chevalier,
  Scott Good, Isabella De~Blasi, Rafik Rhouma, Christopher McMahon, Jean-Paul
  Lam, Thomas Lo, and Christopher~W. Smith.
\newblock A dataset of simulated patient-physician medical interviews with a
  focus on respiratory cases.
\newblock \emph{Scientific Data}, 9\penalty0 (1), June 2022.
\newblock ISSN 2052-4463.
\newblock \doi{10.1038/s41597-022-01423-1}.
\newblock URL \url{http://dx.doi.org/10.1038/s41597-022-01423-1}.

\bibitem[Kim et~al.(2024)Kim, Hahm, and Lee]{kim-etal-2024-generalizing}
Woojin Kim, Sungeun Hahm, and Jaejin Lee.
\newblock Generalizing clinical de-identification models by privacy-safe data
  augmentation using {GPT}-4.
\newblock In Yaser Al-Onaizan, Mohit Bansal, and Yun-Nung Chen, editors,
  \emph{Proceedings of the 2024 Conference on Empirical Methods in Natural
  Language Processing}, pages 21204--21218, Miami, Florida, USA, November 2024.
  Association for Computational Linguistics.
\newblock \doi{10.18653/v1/2024.emnlp-main.1181}.
\newblock URL \url{https://aclanthology.org/2024.emnlp-main.1181/}.

\bibitem[Kuzmin et~al.(2026)Kuzmin, Liu, Lee, and
  Chng]{kuzmin2026streamvoiceanonenhancingutilityrealtime}
Nikita Kuzmin, Songting Liu, Kong~Aik Lee, and Eng~Siong Chng.
\newblock Stream-voice-anon: Enhancing utility of real-time speaker
  anonymization via neural audio codec and language models, 2026.
\newblock URL \url{https://arxiv.org/abs/2601.13948}.

\bibitem[Liu(2024)]{liu2024zeroshotvoiceconversiondiffusion}
Songting Liu.
\newblock Zero-shot voice conversion with diffusion transformers, 2024.
\newblock URL \url{https://arxiv.org/abs/2411.09943}.

\bibitem[Liu et~al.(2023)Liu, Iter, Xu, Wang, Xu, and Zhu]{liu-etal-2023-g}
Yang Liu, Dan Iter, Yichong Xu, Shuohang Wang, Ruochen Xu, and Chenguang Zhu.
\newblock {G}-eval: {NLG} evaluation using gpt-4 with better human alignment.
\newblock In Houda Bouamor, Juan Pino, and Kalika Bali, editors,
  \emph{Proceedings of the 2023 Conference on Empirical Methods in Natural
  Language Processing}, pages 2511--2522, Singapore, December 2023. Association
  for Computational Linguistics.
\newblock \doi{10.18653/v1/2023.emnlp-main.153}.
\newblock URL \url{https://aclanthology.org/2023.emnlp-main.153/}.

\bibitem[Munshi(2026)]{openswara2026}
Jay Munshi.
\newblock Open swara: The largest open-source humanized voice library, 2026.
\newblock URL \url{https://github.com/jaymunshi/open-swara}.

\bibitem[Nair et~al.(2023)Nair, Schumacher, and
  Kannan]{nair-etal-2023-generating}
Varun Nair, Elliot Schumacher, and Anitha Kannan.
\newblock Generating medically-accurate summaries of patient-provider dialogue:
  A multi-stage approach using large language models.
\newblock In Tristan Naumann, Asma Ben~Abacha, Steven Bethard, Kirk Roberts,
  and Anna Rumshisky, editors, \emph{Proceedings of the 5th Clinical Natural
  Language Processing Workshop}, pages 200--217, Toronto, Canada, July 2023.
  Association for Computational Linguistics.
\newblock \doi{10.18653/v1/2023.clinicalnlp-1.26}.
\newblock URL \url{https://aclanthology.org/2023.clinicalnlp-1.26/}.

\bibitem[Nun et~al.(2025)Nun, Birot, Guibon, Lapostolle, and Lerner]{Nun_2025}
Aimé Nun, Olivier Birot, Gaël Guibon, Frédéric Lapostolle, and Ivan Lerner.
\newblock Simsamu - a french medical dispatch dialog open dataset.
\newblock \emph{Computer Methods and Programs in Biomedicine}, 268:\penalty0
  108857, August 2025.
\newblock ISSN 0169-2607.
\newblock \doi{10.1016/j.cmpb.2025.108857}.
\newblock URL \url{http://dx.doi.org/10.1016/j.cmpb.2025.108857}.

\bibitem[Panariello et~al.(2024)Panariello, Tomashenko, Wang, Miao, Champion,
  Nourtel, Todisco, Evans, Vincent, and
  Yamagishi]{panariello2024voiceprivacy2022challengeprogress}
Michele Panariello, Natalia Tomashenko, Xin Wang, Xiaoxiao Miao, Pierre
  Champion, Hubert Nourtel, Massimiliano Todisco, Nicholas Evans, Emmanuel
  Vincent, and Junichi Yamagishi.
\newblock The voiceprivacy 2022 challenge: Progress and perspectives in voice
  anonymisation, 2024.
\newblock URL \url{https://arxiv.org/abs/2407.11516}.

\bibitem[Papadopoulos~Korfiatis et~al.(2022)Papadopoulos~Korfiatis, Moramarco,
  Sarac, and Savkov]{papadopoulos-korfiatis-etal-2022-primock57}
Alex Papadopoulos~Korfiatis, Francesco Moramarco, Radmila Sarac, and Aleksandar
  Savkov.
\newblock {P}ri{M}ock57: A dataset of primary care mock consultations.
\newblock In Smaranda Muresan, Preslav Nakov, and Aline Villavicencio, editors,
  \emph{Proceedings of the 60th Annual Meeting of the Association for
  Computational Linguistics (Volume 2: Short Papers)}, pages 588--598, Dublin,
  Ireland, May 2022. Association for Computational Linguistics.
\newblock \doi{10.18653/v1/2022.acl-short.65}.
\newblock URL \url{https://aclanthology.org/2022.acl-short.65/}.

\bibitem[Park et~al.(2025)Park, Medennikov, Dhawan, Wang, Huang, Koluguri,
  Puvvada, Balam, and Ginsburg]{pmlr-v267-park25h}
Taejin Park, Ivan Medennikov, Kunal Dhawan, Weiqing Wang, He~Huang, Nithin~Rao
  Koluguri, Krishna~C Puvvada, Jagadeesh Balam, and Boris Ginsburg.
\newblock Sortformer: A novel approach for permutation-resolved speaker
  supervision in speech-to-text systems.
\newblock In Aarti Singh, Maryam Fazel, Daniel Hsu, Simon Lacoste-Julien, Felix
  Berkenkamp, Tegan Maharaj, Kiri Wagstaff, and Jerry Zhu, editors,
  \emph{Proceedings of the 42nd International Conference on Machine Learning},
  volume 267 of \emph{Proceedings of Machine Learning Research}, pages
  48153--48169. PMLR, 13--19 Jul 2025.
\newblock URL \url{https://proceedings.mlr.press/v267/park25h.html}.

\bibitem[Peng et~al.(2024)Peng, Huang, Li, Mohamed, and
  Harwath]{peng-etal-2024-voicecraft}
Puyuan Peng, Po-Yao Huang, Shang-Wen Li, Abdelrahman Mohamed, and David
  Harwath.
\newblock {V}oice{C}raft: Zero-shot speech editing and text-to-speech in the
  wild.
\newblock In Lun-Wei Ku, Andre Martins, and Vivek Srikumar, editors,
  \emph{Proceedings of the 62nd Annual Meeting of the Association for
  Computational Linguistics (Volume 1: Long Papers)}, pages 12442--12462,
  Bangkok, Thailand, August 2024. Association for Computational Linguistics.
\newblock \doi{10.18653/v1/2024.acl-long.673}.
\newblock URL \url{https://aclanthology.org/2024.acl-long.673/}.

\bibitem[{Qwen Team}(2025)]{qwen3technicalreport}
{Qwen Team}.
\newblock Qwen3 technical report, 2025.
\newblock URL \url{https://arxiv.org/abs/2505.09388}.

\bibitem[Radford et~al.(2023)Radford, Kim, Xu, Brockman, Mcleavey, and
  Sutskever]{pmlr-v202-radford23a}
Alec Radford, Jong~Wook Kim, Tao Xu, Greg Brockman, Christine Mcleavey, and
  Ilya Sutskever.
\newblock Robust speech recognition via large-scale weak supervision.
\newblock In Andreas Krause, Emma Brunskill, Kyunghyun Cho, Barbara Engelhardt,
  Sivan Sabato, and Jonathan Scarlett, editors, \emph{Proceedings of the 40th
  International Conference on Machine Learning}, volume 202 of
  \emph{Proceedings of Machine Learning Research}, pages 28492--28518. PMLR,
  23--29 Jul 2023.
\newblock URL \url{https://proceedings.mlr.press/v202/radford23a.html}.

\bibitem[Sekoyan et~al.(2025)Sekoyan, Koluguri, Tadevosyan, Zelasko, Bartley,
  Karpov, Balam, and Ginsburg]{sekoyan2025canary1bv2parakeettdt06bv3efficient}
Monica Sekoyan, Nithin~Rao Koluguri, Nune Tadevosyan, Piotr Zelasko, Travis
  Bartley, Nikolay Karpov, Jagadeesh Balam, and Boris Ginsburg.
\newblock Canary-1b-v2 & parakeet-tdt-0.6b-v3: Efficient and high-performance
  models for multilingual asr and ast, 2025.
\newblock URL \url{https://arxiv.org/abs/2509.14128}.

\bibitem[Shen et~al.(2023)Shen, Ju, Tan, Liu, Leng, He, Qin, Zhao, and
  Bian]{shen2023naturalspeech2latentdiffusion}
Kai Shen, Zeqian Ju, Xu~Tan, Yanqing Liu, Yichong Leng, Lei He, Tao Qin, Sheng
  Zhao, and Jiang Bian.
\newblock Naturalspeech 2: Latent diffusion models are natural and zero-shot
  speech and singing synthesizers, 2023.
\newblock URL \url{https://arxiv.org/abs/2304.09116}.

\bibitem[Simancek and Vydiswaran(2024)]{simancek-vydiswaran-2024-handling}
Dalton Simancek and V.G.Vinod Vydiswaran.
\newblock Handling name errors of a {BERT}-based de-identification system:
  Insights from stratified sampling and {M}arkov-based pseudonymization.
\newblock In Elena Volodina, David Alfter, Simon Dobnik, Therese
  Lindstr{\"o}m~Tiedemann, Ricardo Mu{\~n}oz~S{\'a}nchez, Maria~Irena Szawerna,
  and Xuan-Son Vu, editors, \emph{Proceedings of the Workshop on Computational
  Approaches to Language Data Pseudonymization (CALD-pseudo 2024)}, pages 1--7,
  St. Julian{'}s, Malta, March 2024. Association for Computational Linguistics.
\newblock \doi{10.18653/v1/2024.caldpseudo-1.1}.
\newblock URL \url{https://aclanthology.org/2024.caldpseudo-1.1/}.

\bibitem[Singh et~al.(2025)Singh, Dzialo, Kim, Srivatsa, Bulu, Gadde, and
  Kenthapadi]{singh-etal-2025-redactor}
Praphul Singh, Charlotte Dzialo, Jangwon Kim, Sumana Srivatsa, Irfan Bulu, Sri
  Gadde, and Krishnaram Kenthapadi.
\newblock {R}edact{OR}: An {LLM}-powered framework for automatic clinical data
  de-identification.
\newblock In Georg Rehm and Yunyao Li, editors, \emph{Proceedings of the 63rd
  Annual Meeting of the Association for Computational Linguistics (Volume 6:
  Industry Track)}, pages 510--530, Vienna, Austria, July 2025. Association for
  Computational Linguistics.
\newblock ISBN 979-8-89176-288-6.
\newblock \doi{10.18653/v1/2025.acl-industry.36}.
\newblock URL \url{https://aclanthology.org/2025.acl-industry.36/}.

\bibitem[Snyder et~al.(2018)Snyder, Garcia-Romero, Sell, Povey, and
  Khudanpur]{Snyder_2018}
David Snyder, Daniel Garcia-Romero, Gregory Sell, Daniel Povey, and Sanjeev
  Khudanpur.
\newblock X-vectors: Robust dnn embeddings for speaker recognition.
\newblock In \emph{2018 IEEE International Conference on Acoustics, Speech and
  Signal Processing (ICASSP)}, page 5329–5333. IEEE, April 2018.
\newblock \doi{10.1109/icassp.2018.8461375}.
\newblock URL \url{http://dx.doi.org/10.1109/ICASSP.2018.8461375}.

\bibitem[Tan et~al.(2021)Tan, Deng, Yeung, Jiang, Chen, and
  Lee]{tan2021editspeechtextbasedspeech}
Daxin Tan, Liqun Deng, Yu~Ting Yeung, Xin Jiang, Xiao Chen, and Tan Lee.
\newblock Editspeech: A text based speech editing system using partial
  inference and bidirectional fusion, 2021.
\newblock URL \url{https://arxiv.org/abs/2107.01554}.

\bibitem[Tomashenko et~al.(2020)Tomashenko, Srivastava, Wang, Vincent, Nautsch,
  Yamagishi, Evans, Patino, Bonastre, Noé, and
  Todisco]{tomashenko2020introducingvoiceprivacyinitiative}
Natalia Tomashenko, Brij Mohan~Lal Srivastava, Xin Wang, Emmanuel Vincent,
  Andreas Nautsch, Junichi Yamagishi, Nicholas Evans, Jose Patino,
  Jean-François Bonastre, Paul-Gauthier Noé, and Massimiliano Todisco.
\newblock Introducing the voiceprivacy initiative, 2020.
\newblock URL \url{https://arxiv.org/abs/2005.01387}.

\bibitem[Tomashenko et~al.(2024)Tomashenko, Miao, Champion, Meyer, Wang,
  Vincent, Panariello, Evans, Yamagishi, and
  Todisco]{tomashenko2024voiceprivacy2024challengeevaluation}
Natalia Tomashenko, Xiaoxiao Miao, Pierre Champion, Sarina Meyer, Xin Wang,
  Emmanuel Vincent, Michele Panariello, Nicholas Evans, Junichi Yamagishi, and
  Massimiliano Todisco.
\newblock The voiceprivacy 2024 challenge evaluation plan, 2024.
\newblock URL \url{https://arxiv.org/abs/2404.02677}.

\bibitem[Vakili and Dalianis(2022)]{vakili-dalianis-2022-utility}
Thomas Vakili and Hercules Dalianis.
\newblock Utility preservation of clinical text after de-identification.
\newblock In Dina Demner-Fushman, Kevin~Bretonnel Cohen, Sophia Ananiadou, and
  Junichi Tsujii, editors, \emph{Proceedings of the 21st Workshop on Biomedical
  Language Processing}, pages 383--388, Dublin, Ireland, May 2022. Association
  for Computational Linguistics.
\newblock \doi{10.18653/v1/2022.bionlp-1.38}.
\newblock URL \url{https://aclanthology.org/2022.bionlp-1.38/}.

\bibitem[wai Yim et~al.(2023)wai Yim, Fu, Abacha, Snider, Lin, and
  Yetisgen]{yim2023acibenchnovelambientclinical}
Wen wai Yim, Yujuan Fu, Asma~Ben Abacha, Neal Snider, Thomas Lin, and Meliha
  Yetisgen.
\newblock Aci-bench: a novel ambient clinical intelligence dataset for
  benchmarking automatic visit note generation, 2023.
\newblock URL \url{https://arxiv.org/abs/2306.02022}.

\bibitem[Wang et~al.(2022)Wang, Yi, Fu, Tao, and
  Wen]{wang2022campnetcontextawaremaskprediction}
Tao Wang, Jiangyan Yi, Ruibo Fu, Jianhua Tao, and Zhengqi Wen.
\newblock Campnet: Context-aware mask prediction for end-to-end text-based
  speech editing, 2022.
\newblock URL \url{https://arxiv.org/abs/2202.09950}.

\bibitem[Weatherhead et~al.(2026)Weatherhead, Golovko, and
  McCaffrey]{Weatherhead_2026}
James Weatherhead, George Golovko, and Peter McCaffrey.
\newblock Asq-phi: An adversarial synthetic data benchmark for clinical
  de-identification and search utility.
\newblock \emph{Data in Brief}, 65:\penalty0 112586, April 2026.
\newblock ISSN 2352-3409.
\newblock \doi{10.1016/j.dib.2026.112586}.
\newblock URL \url{http://dx.doi.org/10.1016/j.dib.2026.112586}.

\bibitem[Wu et~al.(2024)Wu, Chen, Zhang, Hui, Nezhurina, Berg-Kirkpatrick, and
  Dubnov]{wu2024largescalecontrastivelanguageaudiopretraining}
Yusong Wu, Ke~Chen, Tianyu Zhang, Yuchen Hui, Marianna Nezhurina, Taylor
  Berg-Kirkpatrick, and Shlomo Dubnov.
\newblock Large-scale contrastive language-audio pretraining with feature
  fusion and keyword-to-caption augmentation, 2024.
\newblock URL \url{https://arxiv.org/abs/2211.06687}.

\bibitem[Xiao et~al.(2026)Xiao, Zhao, Ma, Wang, Yu, Liao, and
  Chen]{xiao2026anonymizationeliminationutilitypreservedspeech}
Yunchong Xiao, Yuxiang Zhao, Ziyang Ma, Shuai Wang, Kai Yu, Jiachun Liao, and
  Xie Chen.
\newblock Anonymization, not elimination: Utility-preserved speech
  anonymization, 2026.
\newblock URL \url{https://arxiv.org/abs/2604.17000}.

\bibitem[Zheng et~al.(2023)Zheng, Chiang, Sheng, Zhuang, Wu, Zhuang, Lin, Li,
  Li, Xing, Zhang, Gonzalez, and
  Stoica]{zheng2023judgingllmasajudgemtbenchchatbot}
Lianmin Zheng, Wei-Lin Chiang, Ying Sheng, Siyuan Zhuang, Zhanghao Wu, Yonghao
  Zhuang, Zi~Lin, Zhuohan Li, Dacheng Li, Eric~P. Xing, Hao Zhang, Joseph~E.
  Gonzalez, and Ion Stoica.
\newblock Judging llm-as-a-judge with mt-bench and chatbot arena, 2023.
\newblock URL \url{https://arxiv.org/abs/2306.05685}.

\end{thebibliography}
\end{document}